\documentclass[10pt,twocolumn,letterpaper]{article}

\usepackage[pagenumbers]{wacv}

\usepackage{amsmath,amssymb,amsfonts}
\usepackage{algorithmic}
\usepackage{graphicx}
\usepackage{textcomp}
\usepackage[table]{xcolor}
\definecolor{shadecolor}{rgb}{0.9,0.9,0.9}
\usepackage{framed}
\definecolor{lightgray}{gray}{0.9}
\usepackage{subcaption}
\usepackage{multirow}
\usepackage{siunitx}

\usepackage{booktabs}
\usepackage{threeparttable}

\definecolor{wacvblue}{rgb}{0.21,0.49,0.74}
\usepackage[pagebackref,breaklinks,colorlinks,allcolors=wacvblue]{hyperref}

\def\wacvPaperID{2361} 
\def\confName{WACV}
\def\confYear{2027}

\title{Beyond Representation Learning: A Systematic Study of Joint-Embedding Predictive Generation for 3D Brain MRI}

\author{%
\makebox[\textwidth][c]{%
\begin{tabular}{@{}c@{\hspace{1.2em}}c@{\hspace{1.2em}}c@{\hspace{1.2em}}c@{}}
\textbf{Meng Zhou} &
\textbf{Wenhao You} &
\textbf{Yuxing Chen} &
\textbf{Yueying Tian}
\\
University of Toronto &
University of Waterloo &
University of Alberta &
University of Sussex
\\
{\ttfamily\scriptsize simonzhou@cs.toronto.edu} &
{\ttfamily\scriptsize w22you@uwaterloo.ca} &
{\ttfamily\scriptsize yuxing2@ualberta.ca} &
{\ttfamily\scriptsize yt322@sussex.ac.uk}
\end{tabular}%
}%
}

\begin{document}
\maketitle

\begin{abstract}
Joint-embedding predictive architectures (JEPAs) have primarily been developed for self-supervised representation learning. Denoising JEPA (D-JEPA) recently demonstrated strong generative capabilities on natural images, yet the applicability to 3D medical imaging remains unexplored. Building on the D-JEPA framework, we present \textbf{Med-D-JEPA}, a systematic adaptation and evaluation of joint-embedding predictive generation for 3D brain MRI. Med-D-JEPA operates on continuous latent tokens produced by a 3D KL-regularized adversarial variational autoencoder, and combines masked context prediction, representation-level alignment, per-token diffusion, and iterative next-set-of-token sampling. We evaluate unconditional and class-conditional generation quality on BraTS2019 and OASIS-1 datasets; downstream classification utility; and preliminary whole-tumor segmentation on BraTS2020. Across different generation settings, Med-D-JEPA achieves superior or competitive performance compared to several strong baselines on fidelity and diversity metrics. Compared to training with real samples, Med-D-JEPA-based synthetic pretraining improves classification AUC from 0.63 to 0.85 on BraTS2019 and from 0.78 to 0.87 on OASIS-1. In the segmentation study, pretraining on Med-D-JEPA samples improves Dice from 0.74 to 0.80 and reduces HD95 from 13.40 to 9.56~mm. These findings establish joint-embedding predictive generation as a promising direction for 3D medical image synthesis and encourage further research in this direction. 
\end{abstract}

\section{Introduction}
\label{sec:intro}
\begin{figure}[t]
    \centering
    \includegraphics[width=\linewidth]{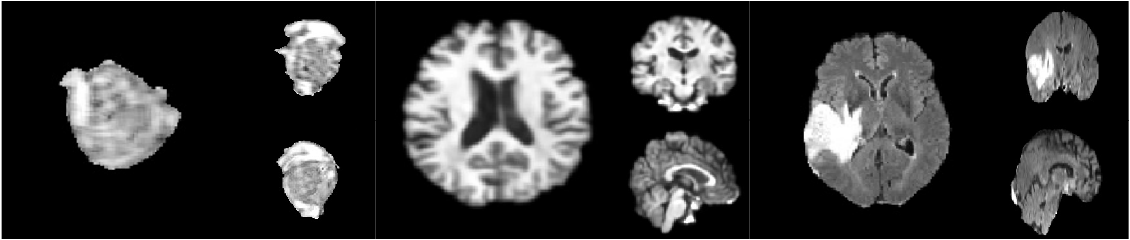}
    \caption{Synthetic MRI volumes generated by Med-D-JEPA. Left to right: orthogonal views of a $128^3$ volume on BraTS2019, a $128^3$ volume on OASIS-1 a
    $256^3$ volume on BraTS2020.}
    \label{fig:start_fig}
\end{figure}
\begin{figure*}[t]
    \centering
    \includegraphics[width=0.99\textwidth]{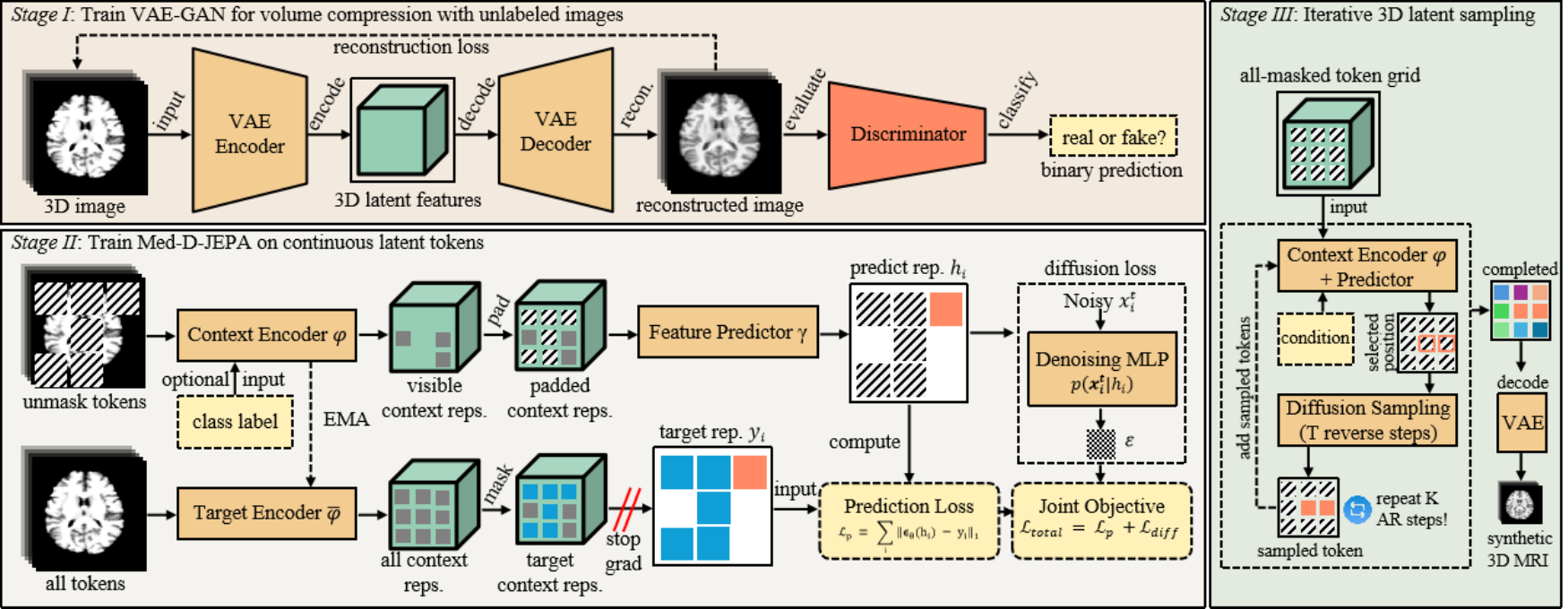}
    \caption{\textbf{Overview of Med-D-JEPA.} A 3D KL-VAE-GAN first compresses MRI volumes into continuous latent tokens. Med-D-JEPA then jointly optimizes a JEPA prediction loss and a per-token diffusion loss, with optional class conditioning. During inference, it progressively samples sets of latent tokens and decodes the completed 3D latent grid into a synthetic MRI volume. ``reps.'' denotes representations, hatched cells indicate masked locations, and orange highlights an example target location.}
    \label{fig:main_archi}
\end{figure*}
3D magnetic resonance imaging (MRI) plays a central role in neuroimaging, supporting anatomical assessment, tumor diagnosis, treatment planning, and longitudinal monitoring~\cite{bakas2017advancing}. However, the development of reliable learning-based systems for volumetric MRI remains constrained by limited cohort sizes, costly expert annotations, heterogeneous acquisition protocols, and restrictions on sharing patient data~\cite{litjens2017survey,yi2019generative}. Synthetic medical image generation has therefore emerged as a promising approach for augmenting scarce datasets, reducing reliance on repeated data acquisition, and supporting the development of data-intensive learning systems~\cite{guo2025maisi}. Nevertheless, visual realism alone does not guarantee that synthetic volumes are useful~\cite{kong2026driving}. Effective medical image generation models must also preserve 3D anatomical coherence, represent relevant pathological characteristics, capture sufficient population diversity, and ultimately provide measurable benefits to downstream clinical tasks. Existing 3D medical image generation models are predominantly based on generative adversarial networks (GANs)~\cite{kwon2019generation,han2018gan,sun2022hierarchical,subramaniam2022generating}, diffusion models~\cite{muller2022diffusion,khader2023denoising,wang20253d,dorjsembe2024conditional,guo2025maisi}, or autoregressive (AR) models~\cite{pinaya2023generative,tudosiu2022morphology,zhou2024conditional,zhou2025generating}. GAN-based approaches enable efficient one-pass volume synthesis, but often suffer from unstable optimization and limited distributional coverage~\cite{gulrajani2017improved}. Diffusion models provide stable training and high sample quality, but even in latent space, 3D generation requires substantial denoising efforts over volumetric representations, resulting in considerable inference latency and nontrivial memory demands~\cite {zhou2025generating}. Autoregressive methods offer an alternative by factorizing the data distribution over a sequence of latent tokens, but they mostly rely on vector-quantized representations, introducing irreversible quantization loss~\cite{li2024autoregressive}. Moreover, conventional token-by-token sampling can become prohibitively slow as the number of 3D latent tokens increases. These trade-offs motivate a framework that combines continuous latent representations with efficient, semantically structured token generation.

Joint-embedding predictive architectures (JEPAs) have recently emerged as an effective paradigm for self-supervised representation learning~\cite{assran2023self,bardes2024revisiting,chen2026vl}. Rather than reconstructing every input detail in pixel space, JEPA predicts the representations of unobserved targets from visible context. Consequently, most existing JEPA research has focused on learning transferable image or video representations instead of generating new samples. Denoising with a Joint-Embedding Predictive Architecture (D-JEPA)~\cite{dengsheng2025denoising} recently bridged these two directions by formulating masked latent prediction as generalized next-token prediction and employing the diffusion objective to model continuous token distributions, and demonstrated strong performance on natural images. Nevertheless, whether this generative formulation remains effective in data-scarce 3D medical imaging has not been systematically investigated.

Extending D-JEPA to volumetric MRI raises several questions specific to its predictive generative formulation. \textbf{First}, can D-JEPA effectively model 3D latent tokens while preserving anatomical and pathological consistency across anatomical planes? \textbf{Second}, can its generative performance be maintained when it is trained on 3D medical imaging datasets that are substantially smaller than the natural-image datasets? \textbf{Third}, does the joint-embedding prediction objective improve 3D image generation, and how does generalized next-set-of-token sampling trade generation quality against inference cost?

Motivated by these challenges, we present \textbf{Med-D-JEPA}, a volumetric adaptation and systematic evaluation of D-JEPA for 3D brain MRI generation. As illustrated in~\cref{fig:start_fig}, Med-D-JEPA supports diverse generation settings spanning tumor-centered ROIs and whole-brain volumes, FLAIR and T1-weighted MRI contrasts, and resolutions from $128^3$ to $256^3$ voxels. Med-D-JEPA compresses each MRI volume into a grid of continuous latent tokens using a 3D KL-regularized adversarial variational autoencoder. We extend the context-target prediction mechanism to 3D latent patches and jointly optimize a JEPA prediction objective and a per-token diffusion objective. During inference, Med-D-JEPA predicts and samples \textit{multiple latent tokens} at each AR step, providing an alternative to both sequential next-token generation and full-volume denoising. We evaluate Med-D-JEPA across \textbf{three complementary brain MRI settings}, assessing both generation quality and downstream utility in classification and segmentation. On \textbf{BraTS2019}, we investigate unconditional and HGG/LGG-conditioned generation and classification on tumor ROIs following prior studies~\cite{zhou2024conditional,zhou2025generating}. On \textbf{OASIS-1}, we evaluate unconditional and cognitive-status-conditioned generation and classification on whole-brain T1-weighted MRI following~\cite{tian2026optimizing}. Finally, on \textbf{BraTS2020}, we investigate unconditional whole-brain MRI generation and conduct a preliminary segmentation study. Our contributions are summarized as follows:

\begin{itemize}
\item We provide, to the best of our knowledge, the first systematic investigation of a D-JEPA-based generative framework for 3D medical imaging, extending continuous joint-embedding prediction from 2D natural images to 3D MRIs.

\item We develop Med-D-JEPA that combines a 3D KL-regularized adversarial variational autoencoder, joint-embedding prediction, and per-token diffusion for unconditional and class-conditional brain MRI generation.

\item We conduct extensive experiments and evaluate Med-D-JEPA across three MRI datasets: BraTS2019, BraTS2020, and OASIS-1. The experiments validate the generalization across \textit{spatial scopes, resolutions, and imaging contrasts}.

\item We show that Med-D-JEPA is a powerful alternative to conventional Diffusion and Autoregressive generation methods based on distributional fidelity and diversity, volumetric consistency, sampling efficiency, and downstream utility.

\end{itemize}

\section{Related Work}
\noindent \textbf{GAN-based Image Synthesis.} GANs were among the earliest approaches to medical image synthesis and have demonstrated the feasibility of generating realistic 2D and 3D medical images~\cite{chartsias2017multimodal,kwon2019generation,han2018gan,xia2020pseudo,subramaniam2022generating,sun2022hierarchical}. The generated lesion images are often used to augment training data for improving downstream tasks to overcome data scarcity issues. Nevertheless, 3D adversarial training remains susceptible to optimization instability, mode collapse, and limited distributional coverage~\cite{gulrajani2017improved}.

\noindent \textbf{Diffusion-based Image Synthesis.} Diffusion models~\cite{ho2020denoising,rombach2022high} have emerged as a prominent alternative to GANs, offering high generative fidelity, stable optimization, and flexible conditioning for medical image synthesis.~\cite{muller2022diffusion}.~\cite{khader2023denoising,dorjsembe2024conditional,peng2024metadata,hamamci2024generatect} demonstrate the effectiveness of DM-based methods in generating high-quality 3D medical images that capture intricate details with minimal artifacts. More recently,~\cite{wang20253d} introduced a Patch-Volume Autoencoder and a bidirectional latent noise estimator for controllable 3D medical image synthesis, whereas MAISI~\cite{guo2025maisi} combined latent diffusion with ControlNet~\cite{zhang2023adding} for high-resolution, segmentation-conditioned CT generation.

\noindent \textbf{Autoregressive-based Image Synthesis.} Autoregressive models have recently attracted considerable attention, partly driven by the success of GPT-style language models~\cite{brown2020language,achiam2023gpt}. When applied to image generation, these models represent images as sequences of visual tokens, typically produced by a VQ-VAE or VQGAN and modeled using a causal or masked Transformer~\cite{razavi2019generating,yu2021vector,esser2021taming,chang2022maskgit,chow2026masked}. In medical imaging,~\cite{zhou2025generating,zhou2024conditional} extended this paradigm to unconditional and class-conditional 3D brain tumor ROI generation. However, discrete tokenization introduces quantization error, which may suppress subtle texture variations, while conventional token-by-token decoding becomes increasingly expensive as the volumetric token sequence grows. Recent studies have challenged the need for discrete visual tokens. MAR~\cite{li2024autoregressive} models continuous latent tokens using a per-token diffusion loss and employs masked autoregressive generation to sample sets of tokens in parallel. However, continuous token generation remains largely unexplored for data-limited 3D medical imaging.

\noindent \textbf{Joint-Embedding Predictive Generation.} JEPA models~\cite{assran2023self,bardes2024revisiting,chen2026vl,chen2024high} have recently emerged as a promising paradigm for self-supervised image and video representation learning. JEPAs learn by predicting target representations from visible context rather than reconstructing pixels. D-JEPA~\cite{dengsheng2025denoising} connected joint-embedding prediction with generation by conditioning diffusion over continuous masked tokens. Existing studies focus on large-scale 2D images, and Med-D-JEPA instead examines whether joint-embedding prediction and next-set-of-token sampling remain effective for 3D brain MRI generation.

\section{Method}
\subsection{Overview}
As shown in~\cref{fig:main_archi}, we present the three stages of Med-D-JEPA. First, we train a 3D KL-regularized adversarial variational autoencoder (VAE-GAN) to compress each MRI volume into continuous latent tokens. Second, we train Med-D-JEPA on the resulting latent tokens using a joint-embedding prediction objective and a per-token diffusion loss~\cite{dengsheng2025denoising}. Finally, generation begins from a fully masked latent grid and progressively samples sets of missing tokens; then the completed grid is decoded by the frozen VAE decoder into a synthetic MRI volume.

\subsection{Continuous 3D Latent Compression}
The compression from the whole MRI volume to continuous 3D latent tokens is done by using a volume compression network. We adapt the VAE-GAN model~\cite{larsen2016autoencoding,rombach2022high} with a 3D PatchGAN discriminator, and trained on combined objectives, which integrate perceptual loss $\mathcal{L}_{\mathrm{lpips}}$~\cite{zhang2018unreasonable}, adversarial loss~\cite{esser2021taming} $\mathcal{L}_{\mathrm{adv}}$, voxel-space $L_1$ reconstruction loss $\mathcal{L}_{\mathrm{recon}}$, and 3D gradient loss~\cite{zhou2025generating} $\mathcal{L}_{\mathrm{grad}}$. These combined objectives ensure that the volume reconstructions adhere closely to the image manifold and enforce local realism and textual details. In addition, we follow~\cite{rombach2022high,guo2025maisi} to add a Kullback-Leibler (KL) regularization $\mathbf{L}_{\mathrm{reg}}$ toward a standard normal on the learned latent features for avoiding high variance latent spaces. 

Given an MRI volume $\mathbf{x} \in \mathbb{R}^{1 \times H \times W \times D}$ in grayscale voxel space, where H denotes the height, W the width, and D the depth, the encoder $E$ of VAE-GAN downsamples x and generates the latent representation $z = E(x) \in \mathbb{R}^{c_z \times h \times w \times d}$ with much smaller spatial dimensions, where $c_z$ denotes the dimension of latent features. The decoder $G$ of KL-VAE-GAN approximates the reconstructed volume $\hat{x} = G(z) = G(E(x))$ from the latent features. A 3D discriminator, denoted as $D_{\mathrm{adv}}$, is utilized to identify and penalize any unrealistic artifacts in the reconstructed volume $\hat{x}$. The overall objective of KL-VAE-GAN to train the volume compression network is defined as follows:
\begin{equation}
\begin{aligned}
\min_{E,G} \max_{D_{adv}}
\Bigl(
    &\lambda_{1}\mathcal{L}_{\mathrm{recon}}\bigl(x, \hat{x}\bigr)
    + \lambda_{2}\mathcal{L}_{\mathrm{lpips}}\bigl(x, \hat{x}\bigr) + \\ &\lambda_{3}\mathcal{L}_{\mathrm{grad}}\bigl(x, \hat{x}\bigr) + \lambda_{4}\mathcal{L}_{\mathrm{reg}}\bigl(E(x)\bigr)
    + \lambda_{5}\mathcal{L}_{\mathrm{adv}}
\Bigr).
\end{aligned}
\end{equation}

Where $\mathcal{L}_{\mathrm{adv}}$ is the GAN objective: $\mathcal{L}_{\mathrm{adv}} = \log D_{\mathrm{adv}}(x) + \log\left(1-D_{adv}\bigl(\hat{x})\bigr)\right)$. $\lambda$ terms are the loss weights for each loss component.

After training, the discriminator is discarded, and the VAE encoder and decoder are frozen. We further partition the latent features $z$ into non-overlapping 3D patches and linearly project them into a sequence, $\mathbb{X}=\{x_i\}_{i=1}^{N}$, where $x_i\in\mathbb{R}^{dp}$ is a continuous latent token, $dp$ is the latent patch dimension, and $N$ is the number of spatial latent patches. The resulting 3D patches are used for Med-D-JEPA training as detailed below.

\subsection{Med-D-JEPA Training}
We follow the exact settings in~\cite{dengsheng2025denoising} to include the generative task training objective along with the conventional JEPA training process to enable the generation capability of JEPA. We use three identical vision transformers as the context encoder $\phi$, target encoder $\bar{\phi}$, and feature predictor $\psi$. The parameters of $\phi$ and $\psi$ are randomly initialized. The parameters of $\bar{\phi}$ are initialized identically to $\phi$ and then updated \textbf{only through} the exponentially moving average with $\alpha = 0.999$. During training, we randomly mask a large subset of the latent tokens and treat the remaining tokens as visible context. The context encoder $\phi$ processes only the visible tokens and captures their
global anatomical and pathological structures. The feature
predictor $\psi$ then uses these context features, together with the spatial positions of the masked tokens, to estimate a contextual representation $\mathbf{h}_i$ for each missing location. Because the tokens originate from a 3D latent grid, the prediction of a missing token can draw information from its surrounding anatomy across all three spatial dimensions, rather than from a single slice. At the same time, the target encoder $\bar{\phi}$ processes the complete, unmasked token sequence. Its output provides a stable target representation for every masked location. A two-layer projection MLP $g(\cdot)$ is used to map each representation of the masked token predicted by the feature predictor toward its corresponding target representation $\mathbf{y}_i$ from $\bar{\phi}$. We apply the $L_1$ prediction loss (
\refstepcounter{equation}\label{jepa_loss}
$\mathcal{L}_{\mathrm{pred}} = \frac{1}{|\mathcal{M}|}
\sum_{i\in\mathcal{M}}
\left|g(\mathbf{h}_i)-\mathbf{y}_i\right|_1$) only at the masked locations $\mathcal{M}$, encouraging the predicted features to capture the high-level anatomical and pathological representations associated with the masked tokens.

\noindent \textbf{Per-token Diffusion Learning.} Following D-JEPA~\cite{dengsheng2025denoising}, Med-D-JEPA focuses on learning the conditional distribution $p(\mathbf{x}_i\mid\mathbf{h}_i)$ for each token rather than modeling the distribution of the complete latent volume. Here, $\mathbf{x}_i$ denotes the $i$-th latent patch, whereas
$\mathbf{h}_i$ is the contextual representation for the masked token predicted by the feature predictor from the visible ones. For denoising, we employ a compact denoising MLP,  $\epsilon_{\theta}(\cdot)$, that consists of residual blocks, LayerNorm~\cite{ba2016layer}, and SiLU~\cite{elfwing2018sigmoid} activations for each masked token. During training, Gaussian noise is added to the original latent token at a randomly sampled diffusion timestep. The diffusion loss objective is defined as:
\begin{equation}
\mathcal{L}_{\mathrm{diff}}
=
\frac{1}{|\mathcal{M}|}
\sum_{i\in\mathcal{M}}
\mathbb{E}_{t,\boldsymbol{\epsilon}_i}
\left[
\left\|
\boldsymbol{\epsilon}_i
-
\epsilon_{\theta}
\left(
\mathbf{x}_i^{t}\mid
t,
\mathbf{h}_i
\right)
\right\|_2^2
\right],
\label{eq:token_diffusion}
\end{equation}

Where $\epsilon_i\sim\mathcal{N}(\mathbf{0},\mathbf{I})$. The total training objective is then formulated as: $\mathcal{L}_{\mathrm{total}} = \mathcal{L}_\mathrm{pred} + \mathcal{L}_{\mathrm{diff}}$

\subsection{Iterative 3D Latent Generation}

At inference time, Med-D-JEPA begins with a fully masked 3D latent grid
and progressively generates sets of missing tokens following~\cite{dengsheng2025denoising}. At each outer
iteration, the context encoder processes all previously sampled tokens,
and the feature predictor estimates contextual representations for the
remaining masked locations. A cosine schedule determines the number of
tokens to be generated at the current iteration, while their spatial
locations are randomly selected.

For each selected location $i$ in step $k_i$, token sampling starts from Gaussian
noise,
$\mathbf{x}_i^{T_{\mathrm{diff}}}\sim
\mathcal{N}(\mathbf{0},\mathbf{I})$.
Conditioned on the predicted contextual representation $\mathbf{h}_i$,
the token is progressively denoised through the reverse diffusion
process~\cite{ho2020denoising}:
\begin{equation}
\mathbf{x}_i^{t-1}
=
\frac{1}{\sqrt{\alpha_t}}
\left(
\mathbf{x}_i^{t}
-
\frac{1-\alpha_t}
{\sqrt{1-\bar{\alpha}_t}}
\epsilon_{\theta}
\left(
\mathbf{x}_i^{t}
\mid
t,\mathbf{h}_i
\right)
\right)
+
\sigma_t\boldsymbol{\delta}_t,
\label{eq:reverse_token_diffusion}
\end{equation}
where
$\boldsymbol{\delta}_t\sim\mathcal{N}(\mathbf{0},\mathbf{I})$,
$\alpha_t$ and $\bar{\alpha}_t$ are determined by the diffusion noise
schedule, and $\sigma_t$ controls the noise variance at timestep $t$.
Starting from $t=T_{\mathrm{diff}}$ and repeatedly applying
Eq.~\eqref{eq:reverse_token_diffusion} produces
$\mathbf{x}_i^{0}$ as a sample from the learned conditional distribution
$p(\mathbf{x}_i\mid\mathbf{h}_i)$.

Once reverse diffusion has been completed for the tokens selected at
the current iteration, their sampled values are fixed and incorporated
as visible context for the next iteration. Across $K$ autoregressive token-generation
iterations, the cosine schedule progressively reduces the proportion
of masked token locations from 100\% to 0\%. A larger $K$ generates
fewer tokens per iteration, allowing later tokens to condition on a
progressively richer volumetric context, whereas a smaller $K$ samples
more tokens in parallel with fewer context updates. In the extreme
case of $K=N$, one token is generated at each iteration, making the
procedure equivalent to conventional next-token generation. When
$K=1$, all tokens are sampled in parallel through one-step token
generation. After all tokens have been filled, the complete token sequence is rearranged into its original 3D latent grid through inverse
patchification. The resulting latent feature is then decoded by the
frozen KL-VAE-GAN decoder to produce the synthetic MRI volume.
\section{Experiments}
\subsection{Datasets and Implementation Details}
We evaluate Med-D-JEPA on three public brain MRI datasets. \textbf{BraTS2019} contains FLAIR MRI from 259 HGG and 76 LGG patients~\cite{bakas2017advancing,bakas2018identifying,menze2014multimodal}. Following~\cite{zhou2024conditional,zhou2025generating}, we extract tumor ROIs and resample them to $128^3$ voxels for unconditional and HGG/LGG-conditioned generation, as well as for the downstream classification task. \textbf{OASIS-1} contains T1-weighted MRI from 416 patients~\cite{marcus2007open}. Following~\cite{tian2026optimizing}, patients with $\mathrm{CDR}=0$ are assigned to the cognitively normal group (CN), while those with $\mathrm{CDR}\in \{0.5,1,2\}$ are assigned to the cognitively impaired group (AD). Volumes are resampled to $128^3$ for unconditional and cognitive-status-conditioned generation, as well as for the downstream classification task. \textbf{BraTS2020}~\cite{bakas2017advancing,bakas2018identifying,menze2014multimodal} contains whole-brain FLAIR MRI and segmentation masks from 369 patients. Volumes are resampled to $256^3$ for unconditional generation and downstream whole-tumor segmentation study. All intensities are normalized to $[0,1]$. We hold out 25 HGG and 25 LGG patients from BraTS2019, 30 CN and 30 AD patients from OASIS-1, and 74 patients from BraTS2020 (according to an 80/20 train-test split). These test patients are excluded from 3D KL-VAE-GAN training, Med-D-JEPA training, and all downstream pretraining and fine-tuning. Separate 3D KL-VAE-GAN and Med-D-JEPA models are trained for each dataset.
\textbf{Evaluation: }For a fair comparison, we generate 300 synthetic volumes per class using each generative method for all experiments. For downstream tasks, a 3D ResNet-50~\cite{hara2017learning} is first pretrained on synthetic volumes and then fine-tuned on a fixed real-data cohort. For segmentation, we apply the same pretraining--fine-tuning protocol to a 3D nnUNet~\cite{isensee2021nnu}, using synthetic image--pseudo-mask pairs for pretraining and real image--mask pairs for fine-tuning. Additional details on the training configurations and pseudo-mask generation procedures are provided in Supplementary Sec.~A and D.


\subsection{Baseline Model and Comparison}
For comparison of image generation results, we consider five state-of-the-art methods, 3D-HA-GAN~\cite{sun2022hierarchical}, Medical Diffusion~\cite{khader2023denoising}, 3D-MedDiffusion~\cite{wang20253d}, 3D-VQGAN~\cite{zhou2025generating} and 3D-VQGAN-cond~\cite{zhou2024conditional}. All baselines are rerun using the same datasets and splits. For classification and segmentation, we establish a baseline where we only use real samples (with traditional augmentations for classification), ensuring a fair comparison with other methods. We evaluate the quality of generated MRI using commonly used metrics~\cite{kwon2019generation,peng2024metadata,zhou2025generating,guo2025maisi}: maximum mean discrepancy (MMD)~\cite{gretton2012kernel}, multi-slice structure similarity (MS-SSIM)~\cite{rosca2017variational}, and the Fr\'echet Inception Distance (FID)~\cite{heusel2017gans}. We compute the FID score in three views (Axial, Coronal, Sagittal) to reflect the nature of medical images; we also include FID-Avg., the average FID across the three views. We further include Precision and Recall~\cite{kynkaanniemi2019improved} to separately characterize the fidelity and coverage of the generated distribution. Classification is evaluated using AUC, F1-score, and Accuracy, while segmentation is evaluated using Dice and HD95.

\section{Results}
\subsection{How Well Does Med-D-JEPA Generate Tumor ROIs?}


\begin{figure}[t]
    \centering
    \includegraphics[width=\linewidth]{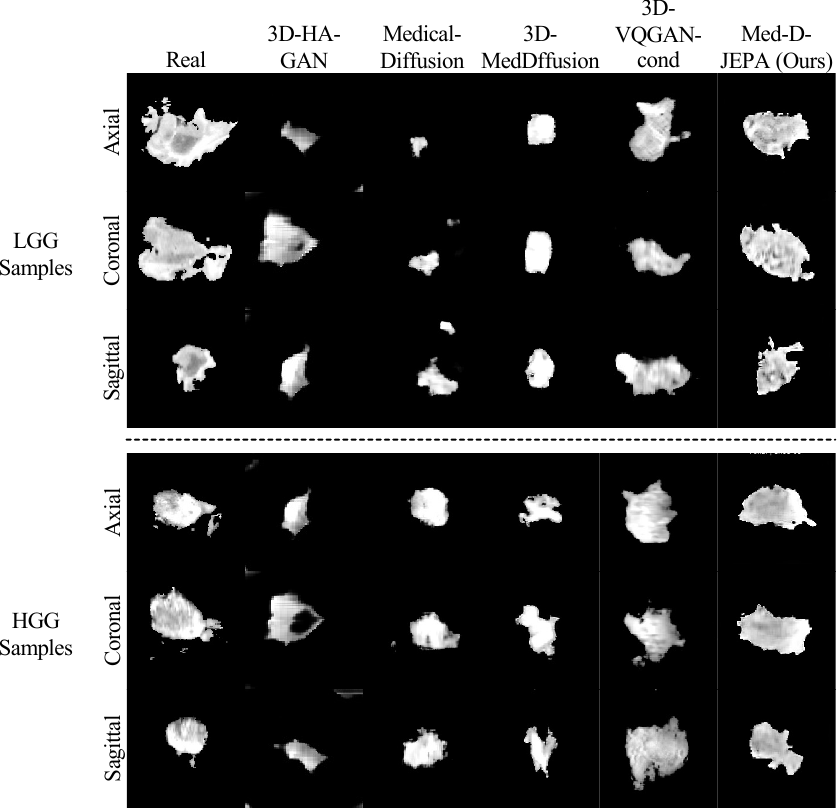}
    \caption{Qualitative comparison of real samples and samples generated by different methods on the BraTS2019 dataset. For each 3D volume, we show the central axial, coronal, and sagittal slices. Best viewed zoomed in.}
    \label{fig:brats19_res}
\end{figure}

In~\cref{fig:brats19_res}, we qualitatively compare real ROIs with LGG- and HGG-conditioned samples generated by the baseline models and Med-D-JEPA. For each 3D volume, we show the central axial, coronal, and sagittal slices. In these examples, 3D-HA-GAN produces ROIs with limited structural detail and noticeable artifacts, while Medical Diffusion exhibits fragmented lesion appearances and noisy boundaries. Samples from 3D-MedDiffusion show unnatural intensity contrast, diminished intratumoral detail and less coherent morphology. Although 3D-VQGAN-cond better preserves contrast and structural detail than the other baselines, fine intratumoral patterns remain blurred. In comparison, Med-D-JEPA produces ROIs with better-defined tumor shapes, internal appearances, intensity contrast, and boundaries, resulting in greater visual similarity to the real examples across all three anatomical planes. More visualization is provided in Supplementary Sec.~B. Quantitatively, as shown in~\cref{tab:main_results_roi}, we report MMD (scaled by $10^{4}$), MS-SSIM, FID-Avg., precision, and recall. Med-D-JEPA achieves competitive performance over other baseline methods on all metrics for unconditional generation, and most of the metrics on class-conditional generation. It has consistently lower MMD and FID-Avg. scores indicate closer agreement between the generated and real-data distributions. Lower MS-SSIM indicates that Med-D-JEPA maintains substantial sample diversity while preserving realistic structural variation. The results for precision and recall suggest that Med-D-JEPA improves sample fidelity without substantially sacrificing coverage of the real-data distribution. More results and analysis are provided in Supplementary Sec.~C.

\begin{table}[t]
\centering
\setlength{\tabcolsep}{2.5pt}

\resizebox{\columnwidth}{!}{%
\begin{tabular}{@{}llccccc@{}}
\toprule
Type & Method
& MMD $\downarrow$
& MS-SSIM (\%)
& FID-Avg. $\downarrow$
& Prec. $\uparrow$
& Rec. $\uparrow$ \\
\midrule

\multicolumn{7}{@{}l}{\textbf{(a) Unconditional}}\\
\addlinespace[1pt]

\multirow{5}{*}{Mixed}
& 3D-HA-GAN~\cite{sun2022hierarchical}
& 2.08 & 95.7{\scriptsize $(+9.0)$} & 90.0
& 0.23$\pm$0.03 & 0.20$\pm$0.05 \\

& Medical Diffusion~\cite{khader2023denoising}
& 1.79 & 90.9{\scriptsize $(+4.2)$} & 33.4
& 0.45$\pm$0.02 & 0.42$\pm$0.04 \\

& 3D-MedDiffusion~\cite{wang20253d}
& \colorbox[HTML]{BFDFBF}{1.76}
& 89.3{\scriptsize $(+2.6)$} & 33.9
& 0.45$\pm$0.04 & 0.50$\pm$0.02 \\

& 3D-VQGAN~\cite{zhou2025generating}
& 1.78
& \colorbox[HTML]{BFDFBF}{88.1{\scriptsize $(+1.4)$}}
& \colorbox[HTML]{BFDFBF}{32.9}
& \colorbox[HTML]{BFDFBF}{0.47$\pm$0.06}
& \colorbox[HTML]{BFDFBF}{0.53$\pm$0.09} \\

& Med-D-JEPA (Ours)
& \colorbox[HTML]{C6DDEC}{1.68}
& \colorbox[HTML]{C6DDEC}{85.4{\scriptsize $(-1.3)$}}
& \colorbox[HTML]{C6DDEC}{26.5}
& \colorbox[HTML]{C6DDEC}{0.58$\pm$0.03}
& \colorbox[HTML]{C6DDEC}{0.57$\pm$0.07} \\

\midrule
\multicolumn{7}{@{}l}{\textbf{(b) Class-conditional}}\\
\addlinespace[1pt]

\multirow{5}{*}{LGG}
& 3D-HA-GAN~\cite{sun2022hierarchical}
& 2.01 & 96.5{\scriptsize $(+11.2)$} & 86.3
& 0.24$\pm$0.02 & 0.18$\pm$0.01 \\

& Medical Diffusion~\cite{khader2023denoising}
& 1.76 & 90.8{\scriptsize $(+5.5)$} & 34.3
& 0.47$\pm$0.08 & 0.39$\pm$0.04 \\

& 3D-MedDiffusion~\cite{wang20253d}
& 1.82 & 93.9{\scriptsize $(+8.6)$} & 56.5
& \colorbox[HTML]{BFDFBF}{0.48$\pm$0.04}
& \colorbox[HTML]{BFDFBF}{0.52$\pm$0.03} \\

& 3D-VQGAN-cond~\cite{zhou2024conditional}
& \colorbox[HTML]{BFDFBF}{1.73}
& \colorbox[HTML]{BFDFBF}{85.5{\scriptsize $(+0.2)$}}
& \colorbox[HTML]{BFDFBF}{29.9}
& 0.46$\pm$0.03
& \colorbox[HTML]{C6DDEC}{0.54$\pm$0.08} \\

& Med-D-JEPA (Ours)
& \colorbox[HTML]{C6DDEC}{1.64}
& \colorbox[HTML]{C6DDEC}{85.1{\scriptsize $(-0.2)$}}
& \colorbox[HTML]{C6DDEC}{23.9}
& \colorbox[HTML]{C6DDEC}{0.59$\pm$0.06}
& 0.51$\pm$0.05 \\

\cmidrule(lr){1-7}

\multirow{5}{*}{HGG}
& 3D-HA-GAN~\cite{sun2022hierarchical}
& 1.96 & 93.8{\scriptsize $(+5.2)$} & 90.1
& 0.26$\pm$0.02 & 0.14$\pm$0.04 \\

& Medical Diffusion~\cite{khader2023denoising}
& \colorbox[HTML]{BFDFBF}{1.51}
& 90.6{\scriptsize $(+2.0)$} & 30.2
& 0.34$\pm$0.07 & 0.31$\pm$0.05 \\

& 3D-MedDiffusion~\cite{wang20253d}
& 1.55 & 89.2{\scriptsize $(+0.6)$} & 29.5
& \colorbox[HTML]{BFDFBF}{0.42$\pm$0.05}
& 0.38$\pm$0.01 \\

& 3D-VQGAN-cond~\cite{zhou2024conditional}
& 1.52
& \colorbox[HTML]{BFDFBF}{87.1{\scriptsize $(-1.5)$}}
& \colorbox[HTML]{BFDFBF}{26.5}
& \colorbox[HTML]{BFDFBF}{0.42$\pm$0.03}
& \colorbox[HTML]{BFDFBF}{0.49$\pm$0.02} \\

& Med-D-JEPA (Ours)
& \colorbox[HTML]{C6DDEC}{1.34}
& \colorbox[HTML]{C6DDEC}{84.9{\scriptsize $(-3.7)$}}
& \colorbox[HTML]{C6DDEC}{26.1}
& \colorbox[HTML]{C6DDEC}{0.52$\pm$0.01}
& \colorbox[HTML]{C6DDEC}{0.56$\pm$0.03} \\

\bottomrule
\end{tabular}
}
\caption{Quantitative comparison of unconditional and LGG/HGG-conditioned generation on BraTS2019. The best and second-best results within each evaluation setting are shown in \colorbox[HTML]{C6DDEC}{blue} and \colorbox[HTML]{BFDFBF}{green}, respectively. For MS-SSIM, values in () denote the difference from the real-data score: 86.7 for mixed, 85.3 for LGG, and 88.6 for HGG.}
\label{tab:main_results_roi}
\end{table}

\subsection{Does Med-D-JEPA Generalize to Whole-Brain MRI generation?}

\begin{figure}[h]
    \centering
    \includegraphics[width=\linewidth]{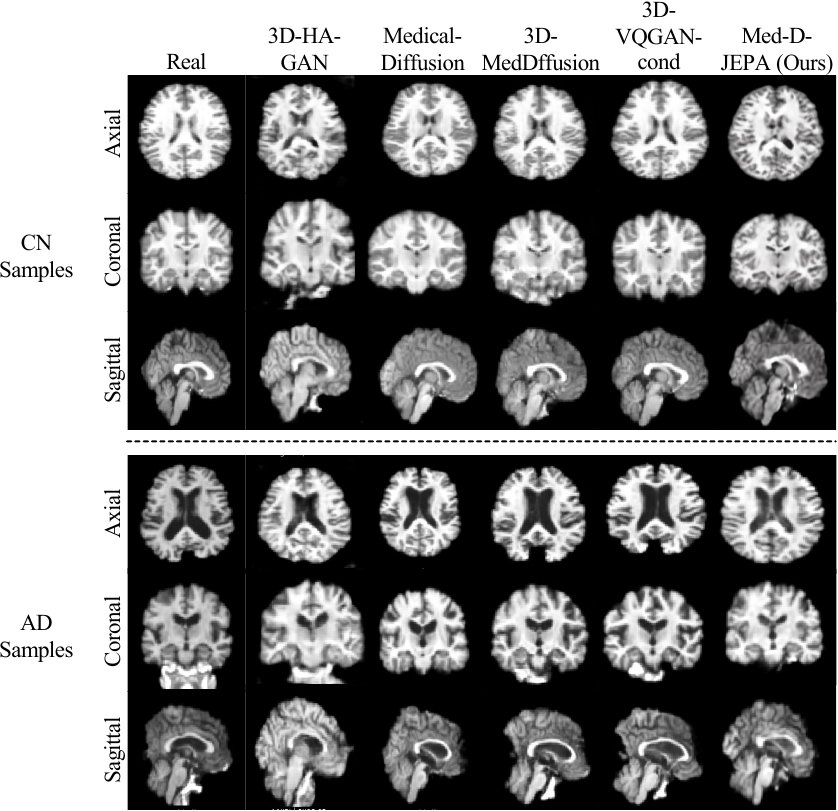}
    \caption{Qualitative comparison of real samples and samples generated by different methods on the OASIS-1 dataset. For each 3D volume, we show the central axial, coronal, and sagittal slices. Best viewed zoomed in.}
    \label{fig:oasis_res}
\end{figure}

In~\cref{fig:oasis_res}, we compare real and CN/AD-conditioned whole-brain T1-weighted generated MRIs, showing the central axial, coronal, and sagittal slices of each volume. Whole-brain generation is more structurally demanding than tumor-ROI synthesis because it requires long-range consistency in ventricular organization, cortical structure, and midline anatomy. The baseline methods generate recognizable brain volumes but exhibit different limitations. 3D-HA-GAN produces relatively coarse structures with less consistent tissue contrast, while Medical Diffusion and 3D-MedDiffusion tend to smooth fine cortical and gray-white matter boundaries. 3D-VQGAN-cond preserves sharper anatomical structures but introduces locally discretized textures. In comparison, Med-D-JEPA better maintains the overall brain contour, bilateral symmetry, ventricular configuration, tissue contrast, and midline structures across all planes. The displayed AD-conditioned sample also exhibits more pronounced ventricular and sulcal enlargement than the CN-conditioned sample, suggesting that the conditioning signal captures group-related structural variation. 

Quantitatively, as shown in~\cref{tab:quali_res_oasis}, Med-D-JEPA demonstrates consistent performance across unconditional and CN/AD-conditioned whole-brain generation. In the unconditional setting, it achieves the lowest MMD and the lowest FID-Avg., while attaining a precision of 0.71 and matching the highest recall of 0.53. For CN-conditioned generation, Med-D-JEPA obtains the best MMD, FID-Avg., and recall, while its precision remains close to the best-performing 3D-VQGAN-cond result (0.76 vs. 0.78). A similar pattern is observed for AD-conditioned generation, together with competitive precision and MS-SSIM. In particular, compared with 3D-VQGAN-cond and Medical Diffusion, Med-D-JEPA maintains comparable precision while improving recall for both CN and AD, suggesting broader coverage without a substantial loss of sample fidelity. Its relatively low MS-SSIM scores indicate increased inter-sample diversity. These results show that Med-D-JEPA generalizes from localized tumor ROIs to whole-brain MRI while preserving both anatomical fidelity and population-level coverage.

\begin{table}[t]
\centering

\setlength{\tabcolsep}{2.5pt}

\resizebox{\columnwidth}{!}{%
\begin{tabular}{@{}llccccc@{}}
\toprule
Cohort & Method
& MMD $\downarrow$
& MS-SSIM (\%)
& FID-Avg. $\downarrow$
& Prec. $\uparrow$
& Rec. $\uparrow$ \\
\midrule

\multicolumn{7}{@{}l}{\textbf{(a) Unconditional}}\\
\addlinespace[1pt]

\multirow{5}{*}{Mixed}
& 3D-HA-GAN~\cite{sun2022hierarchical}
& 1.82 & 82.1{\scriptsize $(+4.9)$} & 72.0
& 0.41$\pm$0.03 & 0.35$\pm$0.07 \\

& Medical Diffusion~\cite{khader2023denoising}
& 1.58 & 80.2{\scriptsize $(+3.0)$} & 27.0
& 0.65$\pm$0.09 & 0.42$\pm$0.03 \\

& 3D-MedDiffusion~\cite{wang20253d}
& 1.60
& \colorbox[HTML]{BFDFBF}{75.2{\scriptsize $(-2.0)$}}
& 29.8
& 0.61$\pm$0.06 & 0.51$\pm$0.05 \\

& 3D-VQGAN~\cite{zhou2025generating}
& \colorbox[HTML]{BFDFBF}{1.50}
& 81.2{\scriptsize $(+4.0)$}
& \colorbox[HTML]{BFDFBF}{26.7}
& \colorbox[HTML]{C6DDEC}{0.72$\pm$0.05}
& \colorbox[HTML]{BFDFBF}{0.53$\pm$0.04} \\

& Med-D-JEPA (Ours)
& \colorbox[HTML]{C6DDEC}{1.46}
& \colorbox[HTML]{C6DDEC}{74.4{\scriptsize $(-2.8)$}}
& \colorbox[HTML]{C6DDEC}{23.6}
& \colorbox[HTML]{BFDFBF}{0.71$\pm$0.07}
& \colorbox[HTML]{C6DDEC}{0.53$\pm$0.01} \\

\midrule
\multicolumn{7}{@{}l}{\textbf{(b) Class-conditional}}\\
\addlinespace[1pt]

\multirow{5}{*}{CN}
& 3D-HA-GAN~\cite{sun2022hierarchical}
& 1.69 & 87.1{\scriptsize $(+6.3)$} & 58.4
& 0.48$\pm$0.01 & 0.30$\pm$0.05 \\

& Medical Diffusion~\cite{khader2023denoising}
& 1.49 & 83.1{\scriptsize $(+2.3)$} & 25.7
& 0.68$\pm$0.02 & 0.40$\pm$0.01 \\

& 3D-MedDiffusion~\cite{wang20253d}
& 1.52
& \colorbox[HTML]{BFDFBF}{79.8{\scriptsize $(-1.0)$}}
& 28.7
& 0.64$\pm$0.03
& \colorbox[HTML]{BFDFBF}{0.52$\pm$0.02} \\

& 3D-VQGAN-cond~\cite{zhou2024conditional}
& \colorbox[HTML]{BFDFBF}{1.41}
& 83.3{\scriptsize $(+2.5)$}
& \colorbox[HTML]{BFDFBF}{25.3}
& \colorbox[HTML]{C6DDEC}{0.78$\pm$0.02}
& 0.50$\pm$0.07 \\

& Med-D-JEPA (Ours)
& \colorbox[HTML]{C6DDEC}{1.37}
& \colorbox[HTML]{C6DDEC}{76.4{\scriptsize $(-4.4)$}}
& \colorbox[HTML]{C6DDEC}{23.1}
& \colorbox[HTML]{BFDFBF}{0.76$\pm$0.04}
& \colorbox[HTML]{C6DDEC}{0.53$\pm$0.03} \\

\cmidrule(lr){1-7}

\multirow{5}{*}{AD}
& 3D-HA-GAN~\cite{sun2022hierarchical}
& 1.63 & 79.6{\scriptsize $(+3.3)$} & 72.9
& 0.27$\pm$0.04 & 0.17$\pm$0.08 \\

& Medical Diffusion~\cite{khader2023denoising}
& 1.56
& \colorbox[HTML]{BFDFBF}{77.5{\scriptsize $(+1.2)$}}
& 25.0
& \colorbox[HTML]{BFDFBF}{0.67$\pm$0.03}
& 0.39$\pm$0.06 \\

& 3D-MedDiffusion~\cite{wang20253d}
& 1.61
& \colorbox[HTML]{C6DDEC}{76.8{\scriptsize $(+0.5)$}}
& 28.4
& 0.55$\pm$0.04
& 0.44$\pm$0.11 \\

& 3D-VQGAN-cond~\cite{zhou2024conditional}
& \colorbox[HTML]{BFDFBF}{1.53}
& 78.1{\scriptsize $(+1.8)$}
& \colorbox[HTML]{BFDFBF}{24.4}
& \colorbox[HTML]{C6DDEC}{0.68$\pm$0.06}
& \colorbox[HTML]{BFDFBF}{0.49$\pm$0.03} \\

& Med-D-JEPA (Ours)
& \colorbox[HTML]{C6DDEC}{1.42}
& 77.6{\scriptsize $(+1.3)$}
& \colorbox[HTML]{C6DDEC}{23.5}
& \colorbox[HTML]{BFDFBF}{0.67$\pm$0.03}
& \colorbox[HTML]{C6DDEC}{0.52$\pm$0.05} \\

\bottomrule
\end{tabular}
}
\caption{
Quantitative comparison of unconditional and CN/AD-conditioned generation on OASIS-1. The best and second-best results within each evaluation setting are shown in \colorbox[HTML]{C6DDEC}{blue} and \colorbox[HTML]{BFDFBF}{green}, respectively. For MS-SSIM, values in () denote the difference from the real-data score: 77.2 for the mixed, 80.8 for CN, and 76.3 for AD.
}
\label{tab:quali_res_oasis}
\end{table}

\subsection{Are Synthetic MRIs Useful for Downstream Classification Tasks?}

\noindent \textbf{On BraTS2019, }we evaluate the downstream utility of synthetic MRI by first pretraining an HGG/LGG classifier on $N_{\mathrm{syn}}$ synthetic samples per class and then fine-tuning it on the real training cohort, as shown in~\cref{tab:classification_brats19}. Synthetic pretraining exposes the classifier to additional anatomical and pathology-related variations, while subsequent real-data fine-tuning adapts the learned representations to the real imaging distribution and mitigates the influence of generation-specific artifacts. Compared with the real-only model, pretraining on 300 Med-D-JEPA samples per class improves AUC, F1-score, and accuracy by 22, 28, and 19 percentage points (pp), respectively. It also outperforms conventional data augmentation by 21, 17, and 16pp on the corresponding metrics. Under the same number of synthetic-data samples, Med-D-JEPA achieves the highest mean performance among all evaluated generative methods, outperforming the strongest baseline, 3D-VQGAN, by 5, 5, and 4pp in AUC, F1-score, and accuracy, respectively. A paired patient-level \textbf{bootstrap analysis with 10,000 resamples} further shows that the improvements over 3D-VQGAN are statistically significant across all three metrics.

\begin{table}[t]
\renewcommand{\arraystretch}{1.0}
\setlength{\tabcolsep}{4.5pt}
\centering
\resizebox{\columnwidth}{!}{
\begin{tabular}{lcccc}
\toprule
Method
& $N_{\mathrm{syn}}$/class
& AUC $\uparrow$
& F1-Score $\uparrow$
& Accuracy $\uparrow$ \\
\midrule

\multicolumn{5}{l}{\textit{Method comparison}} \\

Real only
& 0 & $0.63\pm0.12$ & $0.51\pm0.23$ & $0.55\pm0.08$ \\

Real + traditional augmentation
& 0
& $0.64{\pm}0.02$
& $0.62{\pm}0.02$
& $0.58{\pm}0.04$ \\

3D-HA-GAN \cite{sun2022hierarchical}
& 300
& $0.62{\pm}0.05$
& $0.66{\pm}0.07$
& $0.61{\pm}0.06$ \\

Medical Diffusion \cite{khader2023denoising}
& 300
& $0.77{\pm}0.04$
& $0.67{\pm}0.02$
& $0.69{\pm}0.05$ \\

3D-MedDiffusion \cite{wang20253d}
& 300
& 0.79$\pm$0.03 & 0.68$\pm$0.04 & 0.70$\pm$0.04 \\

3D-VQGAN-cond \cite{zhou2024conditional}
& 300
& \colorbox[HTML]{BFDFBF}{0.80$\pm$0.02}
& \colorbox[HTML]{BFDFBF}{$0.74{\pm}0.02$}
& \colorbox[HTML]{BFDFBF}{$0.70{\pm}0.01$} \\

Med-D-JEPA (Ours)
& 300
& \colorbox[HTML]{C6DDEC}{0.85$\pm$0.04}
& \colorbox[HTML]{C6DDEC}{0.79$\pm$0.06}
& \colorbox[HTML]{C6DDEC}{0.74$\pm$0.05} \\

\midrule
\multicolumn{5}{l}{\textit{Synthetic-data scaling of Med-D-JEPA}} \\

Med-D-JEPA
& 50
& 0.76$\pm$0.03 & 0.68$\pm$0.07 & 0.72$\pm$0.09 \\

Med-D-JEPA
& 100
& 0.80$\pm$0.05 & 0.74$\pm$0.08 & 0.72$\pm$0.03 \\

Med-D-JEPA
& 500
& 0.82$\pm$0.06 & 0.75$\pm$0.10 & 0.71$\pm$0.07 \\

\bottomrule
\end{tabular}
}
\caption{
Classification performance on BraTS2019. The top section shows the comparison of the performance of different methods. The bottom section shows the performance of Med-D-JEPA on different numbers of pretraining samples. Results are reported as mean$\pm$standard deviation over three runs. Best and second-best results are shown in \colorbox[HTML]{C6DDEC}{blue} and \colorbox[HTML]{BFDFBF}{green}, respectively.
}
\label{tab:classification_brats19}
\end{table}

\noindent \textbf{On OASIS-1, }Med-D-JEPA pretraining yields the strongest overall downstream classification performance, as shown in~\cref{tab:classification_oasis}. Compared to the real-only model, it improves AUC, F1-score, and accuracy by 9, 17, and 10pp, respectively. Med-D-JEPA also outperforms the strongest generative baseline, 3D-VQGAN-cond, by 4pp in AUC and 2pp in accuracy, while matching its mean F1-score with slightly higher variability. The bootstrap analysis indicates that the AUC improvement over 3D-VQGAN-cond is statistically significant, whereas the differences in F1-score and accuracy are not. These results suggest that Med-D-JEPA preserves cognitive-status-related information that transfers to real-data classification, and scales well to whole-brain generation.

\begin{table}[t]
\renewcommand{\arraystretch}{1.0}
\setlength{\tabcolsep}{4.5pt}
\centering
\resizebox{\columnwidth}{!}{
\begin{tabular}{lcccc}
\toprule
Method
& $N_{\mathrm{syn}}$/class
& AUC $\uparrow$
& F1-Score $\uparrow$
& Accuracy $\uparrow$ \\
\midrule

\multicolumn{5}{l}{\textit{Method comparison}} \\

Real only
& 0 & 0.78$\pm$0.08 & 0.60$\pm$0.06 & 0.69$\pm$0.03 \\

Real + traditional augmentation
& 0
& 0.79$\pm$0.04
& 0.68$\pm$0.07
& 0.73$\pm$0.05 \\

3D-HA-GAN \cite{sun2022hierarchical}
& 300
& 0.77$\pm$0.06
& 0.68$\pm$0.02
& 0.74$\pm$0.06 \\

Medical Diffusion \cite{khader2023denoising}
& 300
& 0.80$\pm$0.04
& 0.73$\pm$0.09
& 0.75$\pm$0.03 \\

3D-MedDiffusion \cite{wang20253d}
& 300
& 0.81$\pm$0.03 & 0.74$\pm$0.04 & 0.73$\pm$0.03 \\

3D-VQGAN-cond \cite{zhou2024conditional}
& 300
& \colorbox[HTML]{BFDFBF}{0.83$\pm$0.05}
& \colorbox[HTML]{C6DDEC}{0.77$\pm$0.02}
& \colorbox[HTML]{BFDFBF}{0.77$\pm$0.06} \\

Med-D-JEPA (Ours)
& 300
& \colorbox[HTML]{C6DDEC}{0.87$\pm$0.03}
& \colorbox[HTML]{BFDFBF}{0.77$\pm$0.03}
& \colorbox[HTML]{C6DDEC}{0.79$\pm$0.04} \\

\midrule
\multicolumn{5}{l}{\textit{Synthetic-data scaling of Med-D-JEPA}} \\

Med-D-JEPA
& 50
& 0.78$\pm$0.09 & 0.72$\pm$0.04 & 0.71$\pm$0.07 \\

Med-D-JEPA
& 100
& 0.81$\pm$0.06 & 0.75$\pm$0.03 & 0.74$\pm$0.05 \\

Med-D-JEPA
& 500
& 0.84$\pm$0.03 & 0.73$\pm$0.03 & 0.74$\pm$0.02 \\

\bottomrule
\end{tabular}
}
\caption{
Classification performance on OASIS-1. The top section shows the comparison of the performance of different methods. The bottom section shows the performance of Med-D-JEPA on different numbers of pretraining samples. Results are reported as mean$\pm$standard deviation over three runs. Best and second-best results are shown in \colorbox[HTML]{C6DDEC}{blue} and \colorbox[HTML]{BFDFBF}{green}, respectively.
}
\label{tab:classification_oasis}
\end{table}

\noindent \textbf{Synthetic Data Scaling. }We further investigate how downstream performance changes with the amount of Med-D-JEPA-generated samples by varying $N_{\mathrm{syn}}$ from 50 to 500 samples per class. To probe scaling beyond 300 samples, we generate an additional 200 samples per class to form the 500-sample sets. Overall performance improves as $N_{\mathrm{syn}}$ increases from 50 to 300. On BraTS2019, AUC, F1-score, and accuracy increased by 9, 11, and 2pp from 50 samples per class to 300 samples per class, respectively. Similarly, on OASIS-1, the three metrics improved by 9, 5, and 8pp. However, increasing $N_{\mathrm{syn}}$ to 500 does not yield further gains: AUC, F1-score, and accuracy decreased by 3, 4, and 3pp on BraTS2019 and 3, 4, and 5pp on OASIS-1. The consistent peak at 300 samples per class across both datasets indicates a practical saturation point in our setting. One possible explanation is that repeated sampling from a fixed generator increasingly densifies already captured modes rather than expanding coverage of underrepresented task-relevant variations, while also reinforcing generation-specific patterns. Overall, these results identify 300 synthetic samples per class as an effective number in our setting, providing consistent downstream improvements across both datasets without the diminishing returns observed at 500 samples per class.

\subsection{Preliminary study on downstream segmentation task}

\begin{figure}[t]
    \centering
    \includegraphics[width=\linewidth]{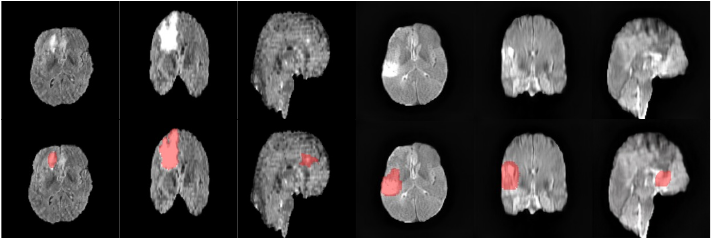}
    \caption{Examples of the NV-Segment-CTMR segmentation results on the synthetic volume.}
    \label{fig:seg_example}
\end{figure}

To complement the classification experiments, we conduct a preliminary evaluation of downstream whole-tumor segmentation on BraTS2020. Specifically, we use NV-Segment-CTMR~\cite{he2025vista3d} to generate pseudo-labels for synthetic volumes produced by Med-D-JEPA. A 3D nnU-Net~\cite{isensee2021nnu} is first pretrained on these synthetic image--mask pairs and then fine-tuned on the real BraTS2020 training set. The real-only baseline uses the same architecture and real-data split but is trained without synthetic pretraining. Further implementation and evaluation details are provided in Supplementary Sec.~D. An example synthetic volume and its corresponding pseudo-label are shown in~\cref{fig:seg_example}. As shown in~\cref{tab:segmentation}, synthetic pretraining increases Dice from $0.74$ to $0.80$ and reduces HD95 from $13.40$ to $9.56$~mm, with both comparisons being statistically significant. Although preliminary, these results provide initial evidence that
Med-D-JEPA-generated volumes are able to support downstream segmentation pretraining and improve performance over the real-only baseline. 

\begin{table}[t]
\renewcommand{\arraystretch}{1.0}
\setlength{\tabcolsep}{5pt}
\centering
\small
\begin{tabular}{lccc}
\toprule
Training strategy
& $N_{\mathrm{syn}}$
& Dice $\uparrow$
& HD95 (mm) $\downarrow$ \\
\midrule
Real only
& 0
& 0.74$\pm$0.03 & 13.40$\pm$5.61 \\



Med-D-JEPA + Real
& 300
& \colorbox[HTML]{C6DDEC}{0.80$\pm$0.06}
& \colorbox[HTML]{C6DDEC}{9.56$\pm$3.97} \\
\bottomrule
\end{tabular}
\caption{Whole-tumor segmentation performance on BraTS2020. Results are reported as mean$\pm$standard deviation over three runs. Best results are shown in \colorbox[HTML]{C6DDEC}{blue}.}
\label{tab:segmentation}
\end{table}

\subsection{Ablation Study}
\noindent \textbf{How Do AR Generation Steps Affect Quality and Efficiency? }As shown in~\cref{tab:abl_ar_steps}, 64 generation steps provide the best overall trade-off, achieving the lowest FID and highest recall on both datasets in approximately 21s per volume. Although 16 steps reduce inference time to about 6s, they provide lower distributional coverage. Conversely, 512 steps (equivalent to the next-token prediction case where $K=N$) yield no quality improvement while being $8.3$--$9.0\times$ slower than 64 steps. We therefore use 64 steps by default, balancing progressive context refinement with parallel token generation.

\begin{table}[h]
\centering
\resizebox{\columnwidth}{!}{
\begin{tabular}{lccccc}
\toprule
Dataset              & Steps                                                   & FID-Avg. $\downarrow$                      & Precision $\uparrow$                     & Recall $\uparrow$                     & Time/Volume $\downarrow$                  \\ \midrule
\multirow{4}{*}{BraTS2019}                                   & 16   & 26.2         &  0.51   & 0.52       & 5.90  \\
                     & 32    &  29.9    &  0.53   & 0.57      &  11.78
                     \\
                     & 64    &  24.6      &   0.57   & 0.58      &  20.60 \\ 
                     & 512 & 28.7 & 0.55 & 0.57 & 185.79 \\ \midrule
\multirow{4}{*}{OASIS-1}            & 16  &  23.3  &  0.65  & 0.51   & 5.88        \\
                     & 32   &   23.2       &  0.66   &  0.55   & 12.71  \\
                     & 64    &   22.6       &   0.65   &   0.57     & 22.10  \\
                     & 512    &   23.0     &  0.64    &  0.57    &  183.50
                     \\ \bottomrule
\end{tabular}
}
\caption{Ablation on the number of autoregressive (AR) sampling steps. Results are based on 100 synthetic samples.}
\label{tab:abl_ar_steps}
\end{table}

\noindent \textbf{Is JEPA Prediction Loss Necessary? }As shown in~\cref{tab:jepa_loss_ablation}, adding $\mathcal{L}_{\mathrm{pred}}$ reduces the average FID by 1.6 and 1.8 on BraTS2019 and OASIS-1, respectively. It also improves precision by 4pp and 5pp and recall by 6pp and 3pp. These consistent gains indicate that representation-level alignment complements per-token diffusion modeling, improving both generation fidelity and distributional coverage.

\begin{table}[h]
\centering
\renewcommand{\arraystretch}{1.0}

\resizebox{\columnwidth}{!}{%
\begin{tabular}{lcccccc}
\toprule
\multirow[c]{2}{*}{Variant}
& \multicolumn{3}{c}{BraTS2019}
& \multicolumn{3}{c}{OASIS-1} \\
\cmidrule(lr){2-4}\cmidrule(lr){5-7}
&
FID-Avg. $\downarrow$
& Prec. $\uparrow$
& Rec. $\uparrow$
& FID-Avg. $\downarrow$
& Prec. $\uparrow$
& Rec. $\uparrow$ \\
\midrule
w/o $\mathcal{L}_{\mathrm{pred}}$
& 28.1 & 0.54 & 0.52
& 25.2 & 0.65 & 0.50 \\

Med-D-JEPA
& 26.5 & 0.58 & 0.57
& 23.6 & 0.71 & 0.53 \\
\bottomrule
\end{tabular}%
}
\caption{Ablation of the JEPA prediction loss.}
\label{tab:jepa_loss_ablation}
\end{table}

\noindent\textbf{Inference Efficiency.}
Finally, we compare the inference time of Med-D-JEPA with those of the baseline methods. Med-D-JEPA generates a $128^3$ volume in 21.35 seconds on average, compared with 15.82 seconds for the strongest baseline, 3D-VQGAN-cond. Although Med-D-JEPA incurs a modest inference-time overhead over 3D-VQGAN-cond, it achieves a favorable trade-off between generation quality and computational efficiency. More ablation results are provided in Supplementary Sec.~E.






\section{Conclusion}
We presented Med-D-JEPA, a joint-embedding predictive generation framework that achieves competitive or superior performance to strong state-of-the-art methods for 3D brain MRI generation. Our results highlight predictive generation as a promising alternative to existing generative paradigms and encourage further research in this direction. Future work will evaluate Med-D-JEPA across additional organs and datasets and extend it to text-conditioned medical image generation.

{
    \small
    \bibliographystyle{ieeenat_fullname}
    \bibliography{main}
}

\newpage
\appendix
\textbf{This supplementary material} is organized as follows: Sec. A provides more details about the datasets and details for model training. Sec. B provides more qualitative results on the Med-D-JEPA visualizations and t-SNE analysis. Sec. C contains additional quantitative results of the tables in the main text, as well as the analysis of the VAE reconstruction ability. Sec. D provides more details on the segmentation study, and Sec. E provides more details on the ablation studies.

\section{Datasets and Implementation Details}
We conduct experiments on three public brain MRI datasets:
BraTS2019 and BraTS2020~\cite{bakas2017advancing,
bakas2018identifying,menze2014multimodal}, and
OASIS-1~\cite{marcus2007open}. These datasets provide complementary
evaluation settings spanning tumor-centered and whole-brain generation,
FLAIR and T1-weighted MRI, 128$^3$ and 256$^3$ resolution, and downstream classification and segmentation tasks.

\noindent \textbf{BraTS2019. }The BraTS2019 dataset contains 259 high-grade glioma (HGG) patients and 76 low-grade glioma (LGG) patients; We use the FLAIR sequence and extract tumor ROIs and resample them to $128\times128\times128$ voxels following~\cite{zhou2024conditional,
zhou2025generating}. Specifically, we first determine the
tumor-containing spatial extent from the nonzero segmentation mask
and apply the same spatial selection to both the FLAIR image and its corresponding mask. We then multiply the FLAIR image by the binary tumor mask and extract a fixed-size crop centered on the tumor region. Voxel intensities are further normalized to the range $[0,1]$. BraTS2019 is used to evaluate
unconditional and HGG/LGG class-conditional generation, as well as the utility of synthetic data for downstream tumor-grade classification.

\noindent \textbf{OASIS-1.} OASIS-1 contains cross-sectional T1-weighted MRI from 416 patients. We define two cognitive-status groups according to the Clinical Dementia Rating (CDR): patients with $\mathrm{CDR}=0$ are assigned to the cognitively normal group (CN), whereas those with
$\mathrm{CDR}\in\{0.5,1,2\}$ are assigned to the cognitively impaired, dementia/AD group (AD)~\cite{khan2019internet}. Following~\cite{tian2026optimizing}, each whole-brain volume is resampled to $128^3$ voxels using MONAI~\cite{cardoso2022monai}, and its voxel intensities are normalized to $[0,1]$. This dataset is used for unconditional and cognitive-status-conditioned whole-brain generation, as well as the utility of synthetic data for downstream CN/AD classification.

\noindent \textbf{BraTS2020. }The BraTS2020 dataset comprises 369 subjects with corresponding tumor segmentation masks. The released volumes are co-registered to a common anatomical template, interpolated to an isotropic resolution of $1\,\mathrm{mm}^{3}$, and skull-stripped. We use the whole-brain FLAIR volumes, resample them to $256^3$ voxels using MONAI~\cite{cardoso2022monai}, and normalize their voxel intensities to $[0,1]$. BraTS2020 is used to evaluate unconditional whole-brain generation and the utility of synthetic data for downstream whole-tumor segmentation.

\noindent \textbf{Implementation Details. }All models are implemented in PyTorch and MONAI~\cite{cardoso2022monai}. We train separate 3D KL-VAE-GAN and Med-D-JEPA models for each dataset. Unless otherwise specified, we
use $K=64$ as the number of autoregressive generation iterations. For
class-conditional generation, we use classifier-free guidance with a
guidance scale of $2.0$.
For KL-VAE-GAN training, we set the loss weights to
$\lambda_{1}=2.0$,
$\lambda_{2}=0.5$,
$\lambda_{3}=1.0$, and rest of the loss weights to 1.0. The KL-regularization weight is set to $\lambda_{4}=1e^{-7}$. KL regularization is activated after the first 1,000 optimization steps and linearly warmed up over the following 1,000 steps. For Med-D-JEPA training, the prediction
and diffusion losses are equally weighted, with 1.0 for both.
For BraTS2019 and OASIS-1, we train the 3D KL-VAE-GAN for 5,000 epochs
using Adam~\cite{kingma2014adam}, with an initial learning rate of
$1e^{-4}$, cosine decay to $1e^{-6}$, 1,000 learning-rate warm-up
steps, and a batch size of 5. Med-D-JEPA extends the official
D-JEPA architecture~\cite{dengsheng2025denoising} from 2D image
tokens to 3D latent patches. For class-conditional models, the
HGG/LGG or CN/AD label is represented by a learned class embedding.
We train Med-D-JEPA for 3,000 epochs using AdamW, with an initial
learning rate of $3e^{-4}$, cosine decay to $5e^{-6}$,
a weight decay of $0.01$, and a batch size of 8.
For the higher-resolution BraTS2020 volumes, we train the 3D
KL-VAE-GAN for 8,000 epochs using an initial learning rate of
$1e^{-4}$, cosine decay to $5e^{-7}$, a weight decay of $0.02$,
and a batch size of 4. Med-D-JEPA is trained for 5,000 epochs, with
the remaining optimization and generation settings unchanged. The code will be made publicly available upon completion of the review process.

\begin{figure}[h]
    \centering
    \includegraphics[width=\linewidth]{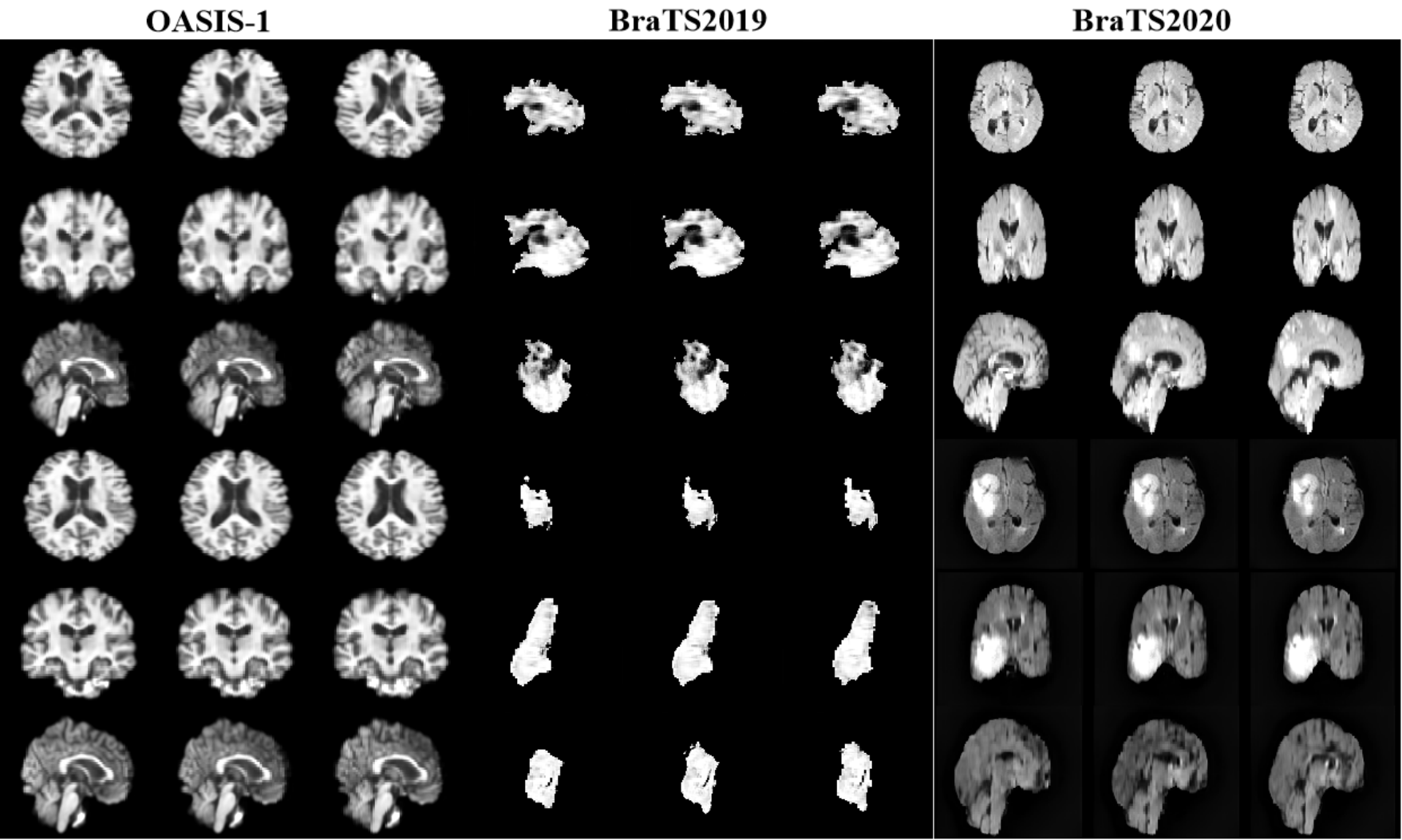}
    \caption{More visualizations on the synthetic volumes from Med-D-JEPA across three datasets.}
    \label{fig:more_vis}
\end{figure}

\noindent \textbf{Evaluation Metrics. }Unless otherwise stated, we generate 300 synthetic volumes for each generative method in the unconditional setting and 300 volumes per class in the class-conditional setting. Following prior
work~\cite{zhou2024conditional,zhou2025generating,guo2025maisi}, we compute the Fréchet Inception Distance (FID)
separately along the axial, coronal, and sagittal planes across the whole volume. Maximum mean discrepancy (MMD) is computed
directly from the complete 3D volumes. For
Precision and Recall, we extract volumetric features using a
MedicalNet~\cite{chen2019med3d} initialized with publicly available pretrained weights.


\section{More on Qualitative Results}

\begin{figure}[h]
    \centering
    \includegraphics[width=\linewidth]{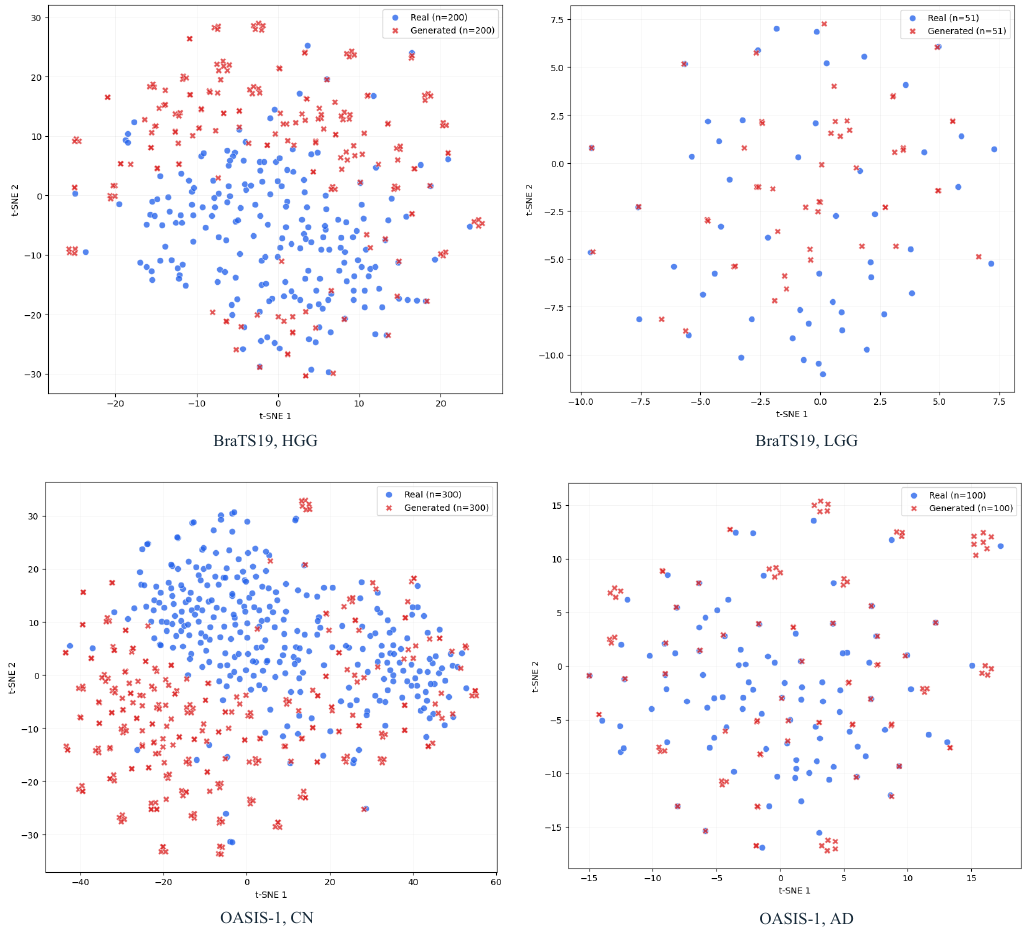}
    \caption{\textbf{TSNE Visualization.} Top: BraTS2019 HGG and LGG results. Bottom: OASIS-1 CN and AD results.}
    \label{fig:tsne}
\end{figure}

We provide additional qualitative examples of synthetic volumes
generated by Med-D-JEPA across the evaluated datasets. For each volume, we display representative axial, coronal, and sagittal slices to assess anatomical consistency across the three spatial planes, as shown in~\cref{fig:more_vis}. We further present t-distributed stochastic neighbor embedding (t-SNE)~\cite{van2008visualizing} visualizations of features extracted from real and synthetic volumes to examine their overlap and local neighborhood structure in the learned feature space.~\cref{fig:tsne} visualizes the difference between real and Med-D-JEPA-generated volumes. The BraTS2019 LGG and OASIS-1 AD cohorts exhibit substantial intermixing between real and generated samples, with no clear separation according to data source. For BraTS2019 HGG and OASIS-1 CN, the generated samples overlap with portions of the real feature space but
also occupy broader, source-dominant regions, indicating a signal of distributional shift. Across all cohorts, the generated samples span multiple regions rather than collapsing into a compact cluster. Overall, the visualization suggests that Med-D-JEPA captures substantial portions of the real feature distributions.

\section{More on Quantitative Results}

\subsection{VAE Reconstruction Ability on Test set}

\begin{figure}[h]
    \centering
    \includegraphics[width=\linewidth]{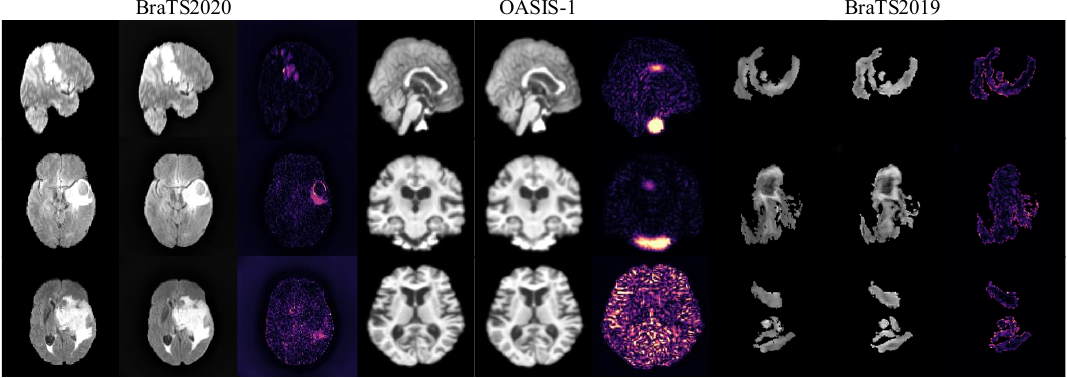}
    \caption{Reconstruction results of the trained 3D KL-VAE-GAN models on the held-out test sets of BraTS2020, OASIS-1, and BraTS2019. For each dataset, the columns show the central slice of the input volume, its reconstruction, and the absolute-error map, respectively.}
    \label{fig:test_recon_res}
\end{figure}

The 3D KL-VAE-GAN determines how much anatomical and pathological
information can be preserved through latent compression and decoding.
Its reconstruction fidelity therefore provides an empirical upper
bound on the image details that can be retained by the subsequent
generative pipeline. We evaluate each trained 3D KL-VAE-GAN on its
corresponding held-out test set and report the peak signal-to-noise
ratio (PSNR) and volumetric structural similarity index measure
(3D-SSIM). Qualitative reconstruction examples and absolute-error maps are presented in~\cref{fig:test_recon_res}. Across all three datasets, the reconstructed volumes preserve the overall anatomy, tumor morphology, and major intensity patterns of the
input volumes. On BraTS2020, the residual errors are primarily
concentrated around tumor boundaries, fine details of the tumor, and regions with heterogeneous signal intensities. On OASIS-1, the global brain structure,
ventricular configuration, and cortical anatomy are well preserved,
while the error maps reveal differences in fine tissue boundaries and
high-frequency textures. Similarly, on the tumor-centered BraTS2019
ROIs, the model preserves the overall lesion shape and internal
intensity distribution, with the largest errors occurring near
irregular boundaries and small disconnected tumor regions. These
results indicate that the learned latent space retains the principal
anatomical and pathological structures required for generation,
although some fine-scale details are attenuated by the compression
process.

As shown in~\cref{tab:vae_reconstruction}, all three models achieve
PSNR values above $30$ dB and 3D-SSIM scores above $0.91$, indicating
that the latent compression stage preserves the principal volumetric
structures of the input MRI. The BraTS2019 model achieves the highest
reconstruction scores, with a PSNR of $33.27$ dB and a 3D-SSIM of
$0.95$, while the two whole-brain models also maintain strong
reconstruction performance at higher spatial and anatomical
complexity.

\begin{table}[t]
\centering
\renewcommand{\arraystretch}{1.0}

\begin{tabular}{llcc}
\toprule
Dataset & $N_\mathrm{test}$ & PSNR & 3D SSIM \\
\midrule
BraTS2020 & 74 & 30.63$\pm$0.89 & 0.91$\pm$0.03 \\
OASIS-1   & 60 & 30.70$\pm$0.99 & 0.92$\pm$0.05 \\
BraTS2019 & 50 & 33.27$\pm$0.10 & 0.95$\pm$0.04 \\
\bottomrule
\end{tabular}
\caption{
Reconstruction performance of the separately trained 3D KL-VAE-GAN
models on their corresponding held-out test sets. Results are reported
as mean$\pm$standard deviation across test volumes.
}
\label{tab:vae_reconstruction}
\end{table}

\subsection{Full Results on Quantitative Comparison}

In this section, we provide the complete quantitative results, including plane-wise FID, for BraTS2019 and OASIS-1 in~\cref{tab:main_results_roi_full} and~\cref{tab:quali_res_oasis_full}. In the unconditional setting, Med-D-JEPA achieves the lowest FID in all three anatomical planes on both datasets. In the class-conditional setting, it obtains the lowest plane-wise FID in nine of the twelve combinations (All 3 for LGG, 2 for HGG, 2 for CN and AD). The exceptions are FID-S for BraTS2019 HGG, FID-S for OASIS-1 CN, and FID-A for OASIS-1 AD.

\begin{table*}[t]
\centering
\setlength{\tabcolsep}{7pt}
\renewcommand{\arraystretch}{1.0}
\resizebox{0.92\textwidth}{!}{
\begin{tabular}{llccccccc}
\toprule
Tumor type
& Method
& MMD ($\times 10^{4}$) $\downarrow$
& MS-SSIM (\%)
& FID-A $\downarrow$
& FID-C $\downarrow$
& FID-S $\downarrow$ 
& Precision $\uparrow$
& Recall $\uparrow$ \\
\midrule

\multicolumn{7}{l}{\textbf{(a) Unconditional generation}} \\
\addlinespace[2pt]

\multirow{5}{*}{Mixed}
& 3D-HA-GAN \cite{sun2022hierarchical}
& 2.08 & 95.7{\scriptsize $(+9.0)$} & 92.6 & 87.8 & 89.5 & 0.23$\pm$0.03 & 0.20$\pm$0.05 \\

& Medical Diffusion \cite{khader2023denoising}
& 1.79 & 90.9{\scriptsize $(+4.2)$} & 33.2 & \colorbox[HTML]{BFDFBF}{31.4} & \colorbox[HTML]{BFDFBF}{35.6} & 0.45$\pm$0.02 & 0.42$\pm$0.04 \\

& 3D-MedDiffusion \cite{wang20253d}
& \colorbox[HTML]{BFDFBF}{1.76} & 89.3{\scriptsize $(+2.6)$} & 30.1 & 34.8 & 36.8 & 0.45$\pm$0.04 & 0.50$\pm$0.02 \\

& 3D-VQGAN \cite{zhou2025generating}
& 1.78 & \colorbox[HTML]{BFDFBF}{88.1{\scriptsize $(+1.4)$}} & \colorbox[HTML]{BFDFBF}{28.5} & 32.3 & 37.9 & \colorbox[HTML]{BFDFBF}{0.47$\pm$0.06} & \colorbox[HTML]{BFDFBF}{0.53$\pm$0.09} \\

& Med-D-JEPA (Ours)
& \colorbox[HTML]{C6DDEC}{1.68} & \colorbox[HTML]{C6DDEC}{85.4{\scriptsize $(-1.3)$}} & \colorbox[HTML]{C6DDEC}{24.8} & \colorbox[HTML]{C6DDEC}{25.7} & \colorbox[HTML]{C6DDEC}{28.9} & \colorbox[HTML]{C6DDEC}{0.58$\pm$0.03} & \colorbox[HTML]{C6DDEC}{0.57$\pm$0.07} \\

\midrule

\multicolumn{7}{l}{\textbf{(b) Class-conditional generation}} \\
\addlinespace[2pt]

\multirow{5}{*}{LGG}
& 3D-HA-GAN \cite{sun2022hierarchical}
& 2.01
& 96.5 {\scriptsize $(+11.2)$}
& 86.2
& 83.5
& 89.3 
& 0.24$\pm$0.02 & 0.18$\pm$0.01 \\

& Medical-Diffusion \cite{khader2023denoising}
& 1.76
& 90.8 {\scriptsize $(+5.5)$}
& 31.7
& 34.2
& 37.1
& 0.47$\pm$0.08 & 0.39$\pm$0.04 \\

& 3D-MedDiffusion \cite{wang20253d}
& 1.82
& 93.9 {\scriptsize $(+8.6)$}
& 43.1
& 57.6
& 68.8 
& \colorbox[HTML]{BFDFBF}{0.48$\pm$0.04} & \colorbox[HTML]{BFDFBF}{0.52$\pm$0.03} \\

& 3D-VQGAN-cond \cite{zhou2024conditional}
& \colorbox[HTML]{BFDFBF}{1.73}
& \colorbox[HTML]{BFDFBF}{85.5 {\scriptsize $(+0.2)$}}
& \colorbox[HTML]{BFDFBF}{25.7}
& \colorbox[HTML]{BFDFBF}{28.9}
& \colorbox[HTML]{BFDFBF}{35.1} 
& 0.46$\pm$0.03 & \colorbox[HTML]{C6DDEC}{0.54$\pm$0.08} \\

& Med-D-JEPA (Ours)
& \colorbox[HTML]{C6DDEC}{1.64}
& \colorbox[HTML]{C6DDEC}{85.1{\scriptsize $(-0.2)$}}
& \colorbox[HTML]{C6DDEC}{21.3}
& \colorbox[HTML]{C6DDEC}{23.4}
& \colorbox[HTML]{C6DDEC}{27.1}
& \colorbox[HTML]{C6DDEC}{0.59$\pm$0.06} & 0.51$\pm$0.05 \\

\cmidrule(lr){1-9}

\multirow{5}{*}{HGG}
& 3D-HA-GAN \cite{sun2022hierarchical}
& 1.96
& 93.8 {\scriptsize $(+5.2)$}
& 90.8
& 85.3
& 94.2 
& 0.26$\pm$0.02 & 0.14$\pm$0.04 \\

& Medical-Diffusion \cite{khader2023denoising}
& \colorbox[HTML]{BFDFBF}{1.51}
& 90.6 {\scriptsize $(+2.0)$}
& 29.8
& 33.5
& \colorbox[HTML]{BFDFBF}{27.4}
& 0.34$\pm$0.07 & 0.31$\pm$0.05 \\

& 3D-MedDiffusion \cite{wang20253d}
& 1.55
& 89.2 {\scriptsize $(+0.6)$}
& 30.3
& 28.6
& 29.7
& 0.42$\pm$0.05 & 0.38$\pm$0.01 \\

& 3D-VQGAN-Cond \cite{zhou2024conditional}
& 1.52
& \colorbox[HTML]{BFDFBF}{87.1 {\scriptsize $(-1.5)$}}
& \colorbox[HTML]{BFDFBF}{28.2}
& \colorbox[HTML]{BFDFBF}{25.5}
& \colorbox[HTML]{C6DDEC}{25.9}
& \colorbox[HTML]{BFDFBF}{0.42$\pm$0.03} & \colorbox[HTML]{BFDFBF}{0.49$\pm$0.02} \\

& Med-D-JEPA (Ours)
& \colorbox[HTML]{C6DDEC}{1.34}
& \colorbox[HTML]{C6DDEC}{84.9{\scriptsize $(-3.7)$}}
& \colorbox[HTML]{C6DDEC}{25.1}
& \colorbox[HTML]{C6DDEC}{25.4}
& 27.9
& \colorbox[HTML]{C6DDEC}{0.52$\pm$0.01} & \colorbox[HTML]{C6DDEC}{0.56$\pm$0.03} \\

\bottomrule
\end{tabular}
}
\caption{
Quantitative comparison of unconditional and class-conditional 3D brain tumor MRI generation on the \textbf{BraTS2019 dataset}. The best and second-best results within each evaluation setting are shown in \colorbox[HTML]{C6DDEC}{blue} and \colorbox[HTML]{BFDFBF}{green}, respectively. For MS-SSIM, values in parentheses denote the signed difference from the corresponding real-data score: 86.7 for mixed, 85.3 for LGG, and 88.6 for HGG.
}
\label{tab:main_results_roi_full}
\end{table*}

\begin{table*}[t]
\centering
\setlength{\tabcolsep}{7pt}
\renewcommand{\arraystretch}{1.0}
\resizebox{0.92\textwidth}{!}{
\begin{tabular}{llccccccc}
\toprule
Cohort type
& Method
& MMD ($\times 10^{4}$) $\downarrow$
& MS-SSIM (\%)
& FID-A $\downarrow$
& FID-C $\downarrow$
& FID-S $\downarrow$ 
& Precision $\uparrow$
& Recall $\uparrow$ \\
\midrule

\multicolumn{7}{l}{\textbf{(a) Unconditional generation}} \\
\addlinespace[2pt]

\multirow{5}{*}{Mixed}
& 3D-HA-GAN \cite{sun2022hierarchical}
& 1.82 & 82.1{\scriptsize $(+4.9)$} & 69.7 & 72.4 & 73.8 & 0.41$\pm$0.03 & 0.35$\pm$0.07 \\

& Medical Diffusion \cite{khader2023denoising}
& 1.58 & 80.2{\scriptsize $(+3.0)$} & \colorbox[HTML]{BFDFBF}{26.3} & 27.8 & 26.9 & 0.65$\pm$0.09 & 0.42$\pm$0.03 \\

& 3D-MedDiffusion \cite{wang20253d}
& 1.60 & \colorbox[HTML]{BFDFBF}{75.2{\scriptsize $(-2.0)$}} & 31.7 & 30.4 & 27.3 & 0.61$\pm$0.06 & 0.51$\pm$0.05 \\

& 3D-VQGAN \cite{zhou2025generating}
& \colorbox[HTML]{BFDFBF}{1.50} & 81.2{\scriptsize $(+4.0)$} & 27.8 & \colorbox[HTML]{BFDFBF}{25.9} & \colorbox[HTML]{BFDFBF}{26.5} & \colorbox[HTML]{C6DDEC}{0.72$\pm$0.05} & \colorbox[HTML]{BFDFBF}{0.53$\pm$0.04} \\

& Med-D-JEPA (Ours)
& \colorbox[HTML]{C6DDEC}{1.46} & \colorbox[HTML]{C6DDEC}{74.4{\scriptsize $(-2.8)$}} & \colorbox[HTML]{C6DDEC}{23.2} & \colorbox[HTML]{C6DDEC}{22.9} & \colorbox[HTML]{C6DDEC}{24.6} & \colorbox[HTML]{BFDFBF}{0.71$\pm$0.07} & \colorbox[HTML]{C6DDEC}{0.53$\pm$0.01} \\

\midrule

\multicolumn{7}{l}{\textbf{(b) Class-conditional generation}} \\
\addlinespace[2pt]

\multirow{5}{*}{CN}
& 3D-HA-GAN \cite{sun2022hierarchical}
& 1.69
& 87.1 {\scriptsize $(+6.3)$}
& 56.9
& 57.6
& 60.8  
& 0.48$\pm$0.01 & 0.30$\pm$0.05 \\

& Medical-Diffusion \cite{khader2023denoising}
& 1.49
& 83.1 {\scriptsize $(+2.3)$}
& 25.1
& 26.6
& 25.4
& 0.68$\pm$0.02 & 0.40$\pm$0.01 \\

& 3D-MedDiffusion \cite{wang20253d}
& 1.52
& \colorbox[HTML]{BFDFBF}{79.8 {\scriptsize $(-1.0)$}}
& 27.8
& 31.5
& 26.7
& 0.64$\pm$0.03 & \colorbox[HTML]{BFDFBF}{0.52$\pm$0.02} \\

& 3D-VQGAN-cond \cite{zhou2024conditional}
& \colorbox[HTML]{BFDFBF}{1.41}
& 83.3{\scriptsize $(+2.5)$}
& \colorbox[HTML]{BFDFBF}{24.5}
& \colorbox[HTML]{BFDFBF}{26.1}
& \colorbox[HTML]{C6DDEC}{25.3}
& \colorbox[HTML]{C6DDEC}{0.78$\pm$0.02} & 0.50$\pm$0.07 \\

& Med-D-JEPA (Ours)
& \colorbox[HTML]{C6DDEC}{1.37}
& \colorbox[HTML]{C6DDEC}{76.4{\scriptsize $(-4.4)$}}
& \colorbox[HTML]{C6DDEC}{22.3}
& \colorbox[HTML]{C6DDEC}{21.3}
& \colorbox[HTML]{BFDFBF}{25.8}
& \colorbox[HTML]{BFDFBF}{0.76$\pm$0.04} & \colorbox[HTML]{C6DDEC}{0.53$\pm$0.03} \\

\cmidrule(lr){1-9}

\multirow{5}{*}{AD}
& 3D-HA-GAN \cite{sun2022hierarchical}
& 1.63
& 79.6 {\scriptsize $(+3.3)$}
& 68.5
& 76.8
& 73.3
& 0.27$\pm$0.04 & 0.17$\pm$0.08 \\

& Medical-Diffusion \cite{khader2023denoising}
& 1.56
& \colorbox[HTML]{BFDFBF}{77.5\scriptsize $(+1.2)$}
& \colorbox[HTML]{BFDFBF}{24.8}
& 25.4
& \colorbox[HTML]{BFDFBF}{24.7}
& \colorbox[HTML]{BFDFBF}{0.67$\pm$0.03} & 0.39$\pm$0.06 \\

& 3D-MedDiffusion \cite{wang20253d}
& 1.61
& \colorbox[HTML]{C6DDEC}{76.8 \scriptsize $(+0.5)$}
& 29.9
& 30.3
& 25.1
& 0.55$\pm$0.04 & 0.44$\pm$0.11 \\

& 3D-VQGAN-Cond \cite{zhou2024conditional}
& \colorbox[HTML]{BFDFBF}{1.53}
& 78.1{\scriptsize $(+1.8)$}
& \colorbox[HTML]{C6DDEC}{24.3}
& \colorbox[HTML]{BFDFBF}{24.1}
& 24.8
& \colorbox[HTML]{C6DDEC}{0.68$\pm$0.06} & \colorbox[HTML]{BFDFBF}{0.49$\pm$0.03}  \\

& Med-D-JEPA (Ours)
& \colorbox[HTML]{C6DDEC}{1.42}
& 77.6{\scriptsize $(+1.3)$}
& \colorbox[HTML]{BFDFBF}{24.8}
& \colorbox[HTML]{C6DDEC}{23.7}
& \colorbox[HTML]{C6DDEC}{21.9}
& \colorbox[HTML]{BFDFBF}{0.67$\pm$0.03} & \colorbox[HTML]{C6DDEC}{0.52$\pm$0.05} \\

\bottomrule
\end{tabular}
}
\caption{
Quantitative comparison of unconditional and class-conditional 3D brain tumor MRI generation on the \textbf{OASIS-1 dataset}. The best and second-best results within each evaluation setting are shown in \colorbox[HTML]{C6DDEC}{blue} and \colorbox[HTML]{BFDFBF}{green}, respectively. For MS-SSIM, values in parentheses denote the signed difference from the corresponding real-data score: 77.2 for the mixed cohort, 80.8 for CN, and 76.3 for AD.}
\label{tab:quali_res_oasis_full}
\end{table*}

\section{More on Segmentation Study}
We use the frozen NV-Segment-CTMR model~\cite{he2025vista3d},
without task-specific fine-tuning, to generate whole-tumor pseudo-masks for the synthetic BraTS2020 volumes. Before applying it to the synthetic data, we evaluate its suitability on the real BraTS2020 \textit{training cohort} using the available ground-truth segmentation masks. NV-Segment-CTMR achieves a Dice score of $0.88\pm0.12$, an IoU of
$0.80\pm0.13$, a precision of $0.91\pm0.12$, a recall of
$0.87\pm0.13$, a specificity of $0.99\pm0.01$, and an HD95 of
$5.01\pm4.17$ mm. Based on this performance, we decide to use the frozen model to predict pseudo-masks for the Med-D-JEPA-generated volumes. These synthetic image--pseudo-mask pairs are subsequently used to pretrain the 3D nnUNet~\cite{isensee2021nnu} before fine-tuning on real image--mask pairs. The segmentation model is pretrained for 100 epochs and subsequently fine-tuned for an additional 100 epochs. Both stages use a batch size of 8
and a constant learning rate of $1e^{-2}$.

\section{More on Ablation Study}

In this section, we provide extended ablation studies on the number of outer autoregressive generation steps and the classifier-free guidance (CFG) scale. The AR-step ablation is conducted using the unconditional
models (with 100 synthetic volumes), with the number of iterations $K$ ranging from 8 to 512.
The CFG ablation is conducted using the class-conditional models, with the generated samples from different classes (100 synthetic volumes per class across both datasets) for evaluation.

As shown in~\cref{tab:abl_qual_res_full_step}, increasing $K$ from 8 to 64 generally improves generation quality and distributional
coverage because later token predictions can condition on a
progressively richer context. On both datasets, $K=64$ achieves the
lowest FID-Avg. and the highest or near-highest Precision and Recall. Further increasing $K$ to 256 or 512 provides no consistent quality improvement while substantially increasing sampling time. In particular, $K=512$ is approximately $9.0\times$ and $8.3\times$ slower than $K=64$ on BraTS2019 and OASIS-1, respectively. We therefore use $K=64$ as the default setting.

\begin{table}[h]
\centering
\resizebox{\columnwidth}{!}{
\begin{tabular}{cccccc}
\hline
Dataset              & Steps                                                   & FID-Avg. $\downarrow$                      & Precision $\uparrow$                     & Recall $\uparrow$                     & Time/Volume $\downarrow$                  \\ \hline
\multirow{4}{*}{BraTS2019}               & 8    & 27.4    &  0.49       & 0.47     & 3.11         \\
                     & 16   & 26.2         &  0.51   & 0.52       & 5.90  \\
                     & 32    &  29.9    &  0.53   & 0.57      &  11.78
                     \\
                     & 64    &  24.6      &   0.57   & 0.58      &  20.60 \\ 
                     & 256 & 29.8 & 0.56 & 0.54 & 92.47 \\
                     & 512 & 28.7 & 0.55 & 0.57 & 185.79 \\ \hline
\multirow{6}{*}{OASIS-1}     
& 8  & 24.6 & 0.63 & 0.49  & 3.13 \\
& 16  &  23.3  &  0.65  & 0.51   & 5.88        \\
                     & 32   &   23.2       &  0.66   &  0.55   & 12.71  \\
                     & 64    &   22.6       &   0.65   &   0.57     & 22.10  \\
                     & 256 & 22.8 & 0.63 & 0.54 & 85.38 \\
                     & 512    &   23.0     &  0.64    &  0.57    &  183.50 \\ \hline
\end{tabular}
}
\caption{Ablation Study on the number of AR steps. using 100 generated volumes.
Sampling time is reported as the average number of seconds per volume.}
\label{tab:abl_qual_res_full_step}
\end{table}

We further investigate the effect of the classifier-free guidance
scale with $K$ fixed at 64. As shown
in~\cref{tab:abl_qual_res_full_cfg}, $s_{\mathrm{cfg}}=2.0$
consistently provides the best overall balance between fidelity and
distributional coverage. On BraTS2019, it achieves the lowest
FID-Avg. and the highest Precision and Recall. On OASIS-1, it also
achieves the lowest FID-Avg. and the highest Precision, while matching the highest Recall. Increasing the guidance scale does not improve FID and generally reduces both Precision and Recall, suggesting that stronger guidance restricts the generated distribution without providing a corresponding gain in fidelity. We therefore use $s_{\mathrm{cfg}}=2.0$ as the default setting.

\begin{table}[h]
\centering
\resizebox{\columnwidth}{!}{
\begin{tabular}{ccccc}
\hline
Dataset              & CFG                                                   & FID-Avg. $\downarrow$                      & Precision $\uparrow$                     & Recall $\uparrow$                        \\ \hline
\multirow{4}{*}{BraTS2019}               & 2.0    & 25.3    &  0.57       & 0.55        \\
                     & 2.5   & 25.8   &  0.54   & 0.52     \\
                     & 3.0    &  25.6   &  0.53   & 0.53 
                     \\
                     & 3.5    &  25.9     &   0.52   & 0.52   \\ 
                     & 4.0 & 25.5 & 0.51 & 0.52 \\ \hline
\multirow{6}{*}{OASIS-1}     
& 2.0  & 23.5 & 0.65 & 0.57  \\
& 2.5  &  23.9  & 0.64  & 0.57        \\
                     & 3.0   &   23.7       &  0.64   &  0.56   \\
                     & 3.5    &   23.6      &   0.63   &   0.55   \\
                     & 4.0 & 23.8 & 0.63 & 0.56 \\ \hline
\end{tabular}
}
\caption{Ablation Study on different cfg values}
\label{tab:abl_qual_res_full_cfg}
\end{table}

\subsection{Sampling Efficiency}
We compare the average sampling time of the baseline
methods in~\cref{tab:sampling_time}. At a resolution of $128^3$, 3D-HA-GAN is the fastest method because it generates each volume in a single forward pass. Med-D-JEPA requires 21.35 seconds per volume, making it slower than Medical Diffusion and 3D-VQGAN-cond because of its iterative next-set-of-token sampling and per-token denoising. Nevertheless, Med-D-JEPA is approximately 22.1\% faster than 3D-MedDiffusion while achieving stronger overall generation quality. At $256^3$ resolution, Med-D-JEPA requires 42.54 seconds per volume. Developing more efficient sampling strategies that reduce the cost of iterative token generation and denoising without compromising generation quality therefore represents an important direction for future work.

\begin{table}[h]
\centering
\begin{tabular*}{\columnwidth}{@{\extracolsep{\fill}}lcc@{}}
\toprule
Method & Resolution & Time/volume (s) $\downarrow$ \\
\midrule
3D-HA-GAN          & $128^3$ & 1.25  \\
Medical Diffusion  & $128^3$ & 8.72  \\
3D-VQGAN-cond      & $128^3$ & 15.82 \\
Med-D-JEPA         & $128^3$ & 21.35 \\
3D-MedDiffusion    & $128^3$ & 27.41 \\
\midrule
Med-D-JEPA         & $256^3$ & 42.54 \\
\bottomrule
\end{tabular*}
\caption{
Average sampling time per volume.
}
\label{tab:sampling_time}
\end{table}

\end{document}